\documentclass[10pt,twocolumn,letterpaper]{article}

\usepackage[pagenumbers]{iccv} 

\definecolor{iccvblue}{rgb}{0.21,0.49,0.74}
\usepackage[pagebackref,breaklinks,colorlinks,allcolors=iccvblue]{hyperref}

\usepackage{multirow}
\usepackage{adjustbox}
\usepackage{amsmath}
\usepackage{array}
\usepackage{graphicx}
\usepackage{amssymb}
\usepackage{color, colortbl}
\usepackage{pifont}
\newcolumntype{M}[1]{>{\centering\arraybackslash}m{#1}}

\newcommand{\cmark}{\ding{51}}%
\newcommand{\xmark}{\ding{55}}%
\definecolor{Gray}{gray}{0.9}

\def\paperID{12534} 
\def\confName{ICCV}
\def\confYear{2025}

\title{Gaussian Splatting is an Effective Data Generator for 3D Object Detection}

\author{
Farhad G. Zanjani \qquad Davide Abati \qquad Auke Wiggers \qquad Dimitris Kalatzis \\ Jens Petersen \qquad Hong Cai \qquad Amirhossein Habibian\\
Qualcomm AI Research\thanks{Qualcomm AI Research is an initiative of Qualcomm Technologies, Inc.}\\
\small{\texttt{\{fzanjani,dabati,auke,dkalatzi,jpeterse,hongcai,ahabibia\}@qti.qualcomm.com}}
}

\begin{document}
\maketitle


\begin{abstract}
We investigate data augmentation for 3D object detection in autonomous driving. We utilize recent advancements in 3D reconstruction based on Gaussian Splatting for 3D object placement in driving scenes. Unlike existing diffusion-based methods that synthesize images conditioned on BEV layouts, our approach places 3D objects directly in the reconstructed 3D space with explicitly imposed geometric transformations. This ensures both the physical plausibility of object placement and highly accurate 3D pose and position annotations.
Our experiments demonstrate that even by integrating a limited number of external 3D objects into real scenes, the augmented data significantly enhances 3D object detection performance and outperforms existing diffusion-based 3D augmentation for object detection. Extensive testing on the nuScenes dataset reveals that imposing high geometric diversity in object placement has a greater impact compared to the appearance diversity of objects. Additionally, we show that generating hard examples, either by maximizing detection loss or imposing high visual occlusion in camera images, does not lead to more efficient 3D data augmentation for camera-based 3D object detection in autonomous driving.

\end{abstract}
\section{Introduction}

The accurate detection of street objects in 3D is a central perception problem in automotive compute vision, and it represents a key requirement which modern assisted and automated driving solutions build on top of.
One of the key challenges in deploying accurate detectors is still the collection of training data.
Indeed, automotive applications involve complicated acquisition pipelines, typically comprising complex setups with multiple sensors and cameras and significant effort for labeling 3D objects~\cite{caesar2020nuscenes,yu2020bdd100k,waymoopen}.

For this reason, the idea to generate synthetic scenes to support the training of 3D perception models is a long standing in automotive~\cite{carla,torcs,deepdriving}, and its appeal only increases as the research community continues to deliver improvements in generative models~\cite{rombach2022stablediffusion,svd,opensora}.

Modern strategies involve generating scenes compliant with a desired 3D layout (\ie, a set of 3D bounding boxes) by means of diffusion models.
To achieve this, the layout is first transformed a control signal~\cite{wu2024neuralassets,bhat2023loosecontrol} or integrated into a text prompt~\cite{gao2023magicdrive}, and it is then fed to the model for scene rendering.
Although this procedure showed great potential in synthesizing realistic scenes, it only involves an \emph{implicit} conditioning on the desired layout, meaning that it offers no guarantees on whether the desired objects will be generated, nor on whether they will appear in the specified location and pose.
The negative impact of such image/layout discrepancies to augmenting detector training remains unclear.
Moreover, diffusion models are still extremely computationally intensive, and their application to automotive settings is currently limited to low resolution scenarios.

\begin{figure}
\centering
 \includegraphics[width=\linewidth, trim={7cm, 4.0cm, 8cm, 3.5cm}, clip]{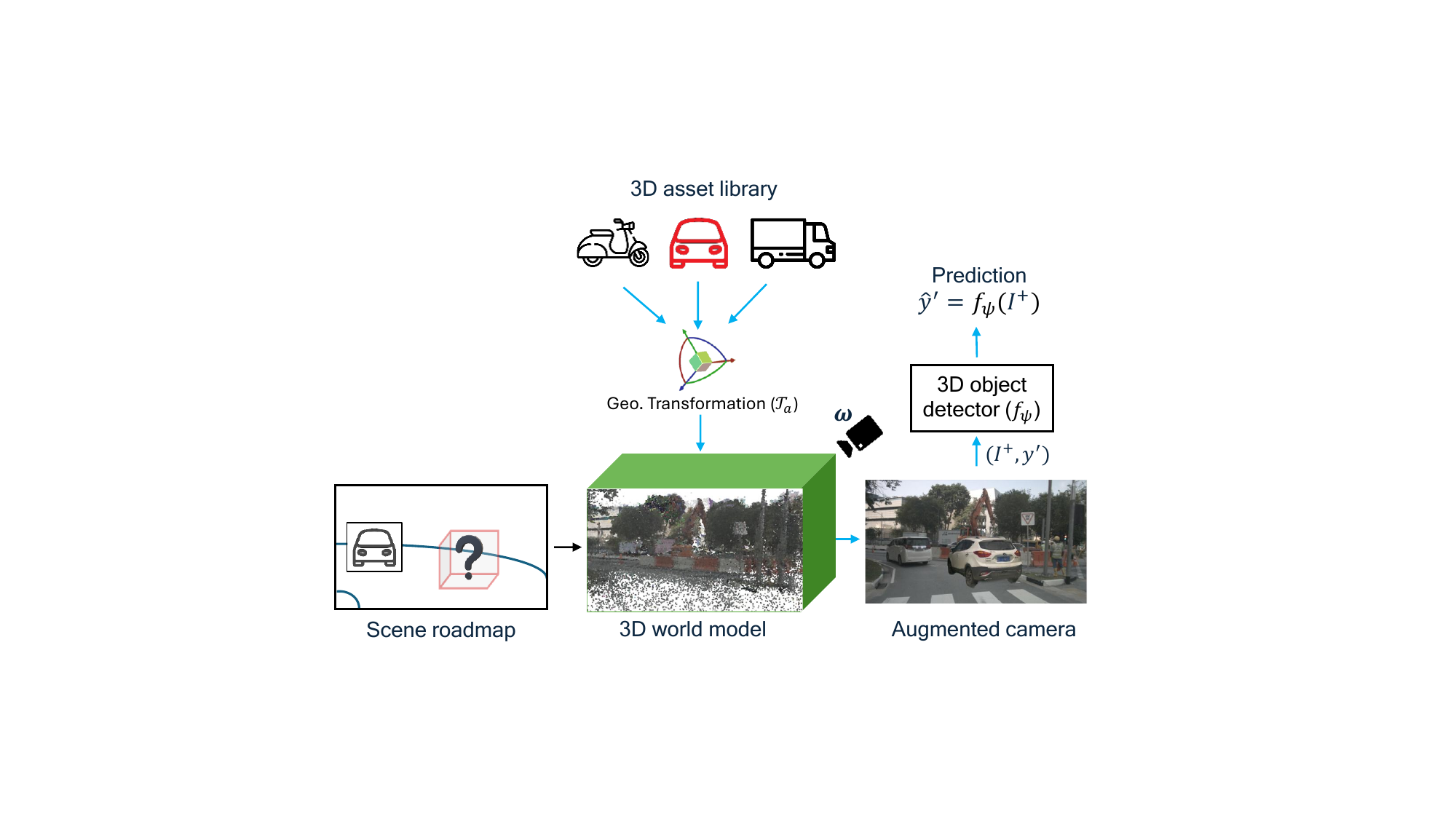}
\vspace{-8pt}
\caption{\small Image augmentation through 3D scene reconstruction and physically plausible object placement for 3D object detection.}
\label{fig:overview}
\vspace{-.5cm}   
\end{figure}

In this work, we therefore focus on augmentation via a 3D reconstruction approach, which allows to maintain \emph{explicit} control over synthetic objects placement, resulting in a high geometric fidelity to their pose and positions in the scene.
More specifically, we explicitly model the scene using a dynamic 3D Gaussian Splatting (3D-GS) approach~\cite{chen2024omnire}, and insert 3D assets into the scene in physically plausible locations by applying explicit geometric transformations.
These synthetically augmented frames will have high location/pose accuracy by design while the photometric diversity is upper-bounded by the size of the 3D asset library.
An overview of our approach is illustrated in \cref{fig:overview}.

We show that the resulting augmented frames boost the performance of 3D detectors in both monocular and multi-view settings.
We further question the importance of appearance diversity by ablating the number of synthetic assets, finding that our solution outperforms diffusion-based baselines even in an extreme setup where a single object per category is used to augment the full dataset.

Finally, we explore some alternative strategies for augmentation including varying the added object's orientation and placing the object in hard locations for the detector.
In this respect, our experiments show that random orientations and locations are more rewarding, possibly because they improve the variations in the data.



\noindent 
Our contributions are as follows:
\begin{itemize}
\item 
   We investigate the potential of advanced 3D driving scene reconstruction to improve 3D object detection performance. To the best of our knowledge, this is the first study to successfully explore 3D detector augmentation using 3D Gaussian Splatting methods.
\item 
  Through extensive experiments, we demonstrate that geometric diversity plays a more crucial role in 3D augmentation than photometric diversity, and that introducing even a limited number of 3D objects can outperform state-of-the-art diffusion-based data augmentation methods.
\end{itemize}

\section{Related work}

\paragraph{3D object detection in street scenes.}
In autonomous driving, 3D object detection is crucial, and approaches differ by sensors and temporal inputs. 
Monocular 3D detection, relying on single camera inputs, offers cost-effective but face challenges in depth perception. FCOS3D~\cite{wang2021fcos3d} is one of the camera-based monocular object detector shows good robustness in compared to other detectors~\cite{xie2301adversarial}. Bird’s-eye-view (BEV) object detection aims at detecting objects in BEV space given multi-view 2D images~\cite{liu2023sparsebev, li2024bevformer,yang2023bevformer}. 
SparseBEV~\cite{liu2023sparsebev} is one of the most efficient solutions to date, consisting of a query-based one stage detector with multiple decoder layers. 
In this paper, we use FCOS3D and a non-temporal configuration of SparseBEV to evaluate the effectiveness of the proposed data augmentation.
\paragraph{3D reconstruction for autonomous driving.}
Recent advancements in NeRF-based~\cite{mildenhall2021nerf} scene reconstruction methods for dynamic autonomous driving, such as SUDS~\cite{turki2023suds} and EmerNeRF~\cite{yang2023emernerf}, have demonstrated impressive reconstruction capabilities. However, these methods model all dynamic elements using a single dynamic field, which limits their controllability and practicality as simulators. 
By decomposing the scene into separate components (represented as nodes in a scene graph), it becomes possible to control them individually. 
This approach is widely adopted in methods like Neural Scene Graphs~\cite{ost2021neural}, UniSim~\cite{yang2023unisim}, MARS~\cite{wu2023mars}, NeuRAD~\cite{tonderski2024neurad}, and ML-NSG~\cite{fischer2024multi}.

Simulating and image rendering of a large-scale autonomous driving dataset with hundreds of scenes using NeRF-based models is computationally heavy and almost impractical. 
Recent 3D Gaussian splatting-based (3D-GS)~\cite{kerbl20233d} works, such as StreetGaussians~\cite{yan2024street}, DrivingGaussians~\cite{zhou2024drivinggaussian}, and HUGS~\cite{zhou2024hugs}, utilize the fast differentiable rendering pipeline of 3D-GS to address the computational hurdle. 
However, these methods typically handle only rigid objects due to the limitations of time-independent representations or deformation-based techniques. 
To address these limitations, OmniRe~\cite{chen2024omnire} proposes a Gaussian scene graph that incorporates various Gaussian representations for both rigid and non-rigid objects. 
This provides additional flexibility and controllability for diverse actors.
In our work, we rely on OmniRe as the scene model to interact with the scene, and we explore its suitability to generate training data for 3D detectors.

\paragraph{Diffusion methods for autonomous driving.}
Diffusion-based models, which model visual context implicitly, demonstrate impressive controllability in image generation across various conditioning scenarios~\cite{gligen}.
This controllability can be leveraged to generate paired images and attributes, such as image labels or annotations like 2D or 3D bounding boxes.
This has proven effective in generating data for 2D object detection~\cite{geodiffusion}. 

Several works extend this idea to 3D object detection tasks in autonomous driving settings.
BEVGen~\cite{swerdlow2024street} conditions image generation on BEV maps for both roads and vehicle bounding boxes. 
The lack of height information in such conditioning has been tackled by BEVControl~\cite{yang2023bevcontrol}. 
A subsequent work, MagicDrive~\cite{gao2023magicdrive}, separates BEV and 3D bounding box conditioning, further enhancing controllability over image generation. 
The authors also demonstrate that the synthesized images can be used to augment training data, improving 3D object detection and segmentation models. A follow-up work, MagicDrive3D~\cite{gao2024magicdrive3d}, combines MagicDrive with deformable GS to modify the generated sequences, albeit only at low resolution. MagicDriveDiT \citep{gao2024magicdrivedit} extends MagicDrive to video data, replacing the convolutional architecture with a Diffusion Transformer~\cite{peebles_dit} and training from scratch on nuScenes instead of starting from a pretrained model.
In these models, 3D control over the desired scene layout remains only implicit, and high resolution generation is troublesome.
We hereby explore augmentation by 3D-GS, that allows explicit 3D control over scene agents and fast high-resolution rendering.

\section{Method}
\begin{figure*}[t!]
    \centering
    \includegraphics[width=0.95\linewidth, trim={1cm, 4cm, 1cm, 4cm}, clip]{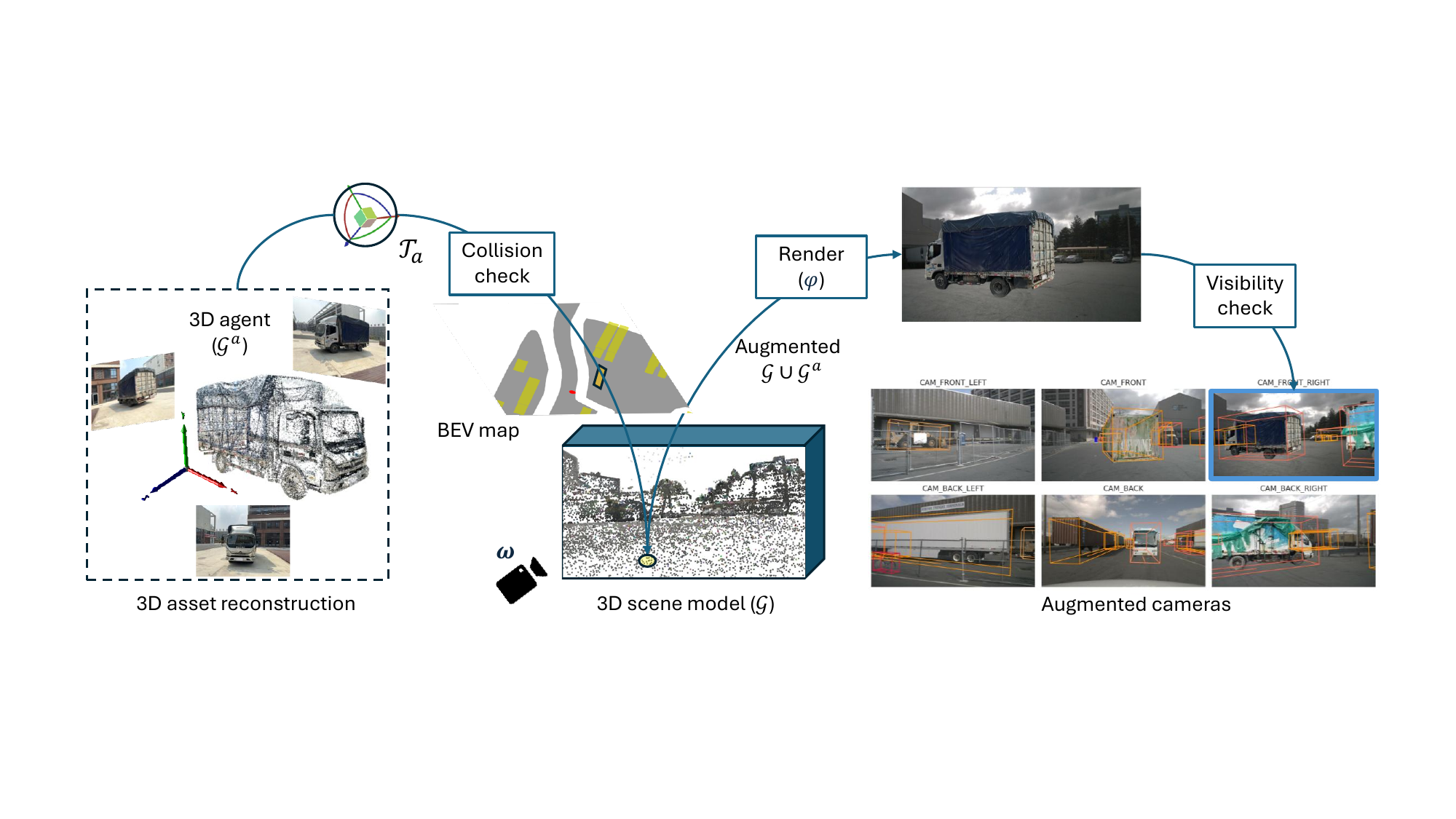}
    \vspace{-15pt}
    \caption{\small Overview of multi-camera 3D data augmentation through object placement in 3D field of Gaussian Splatting.}
    \label{fig:appendix:overview_2}
    \vspace{-.3cm}
\end{figure*}


By processing driving videos along with their associated roadmaps and 3D bounding boxes, we create a 3D world model of the scene using an instance-aware 3D Gaussian Splatting method. 
This 3D model serves as a digital twin of the scene, and enables manipulation of object geometry or introduction of new objects into the scene, thereby allowing the creation of novel scenarios.
We first provide background information on Gaussian Splatting, and then explain how we determine suitable locations for new objects.


\subsection{Preliminaries}\label{sec:preliminaries}

Each scene is modeled by a 3D Gaussian Splatting~\cite{kerbl20233d} (3D-GS) approach, which represents the sequence as a set of 3D Gaussian primitives $g \in \mathcal{G}$, each parametrized by their opacity $o \in (0,1)$, mean position $\mathbf{\mu} \in \mathbb{R}^3$, rotation $\mathbf{q} \in \mathbb{R}^4$ in quaternion, anisotropic scaling factors $\mathbf{s} \in \mathbb{R}^3_{+}$, and view dependent colors $\mathbf{c} \in R^m$ represented through Spherical Harmonics (SH) coefficients. 
Usually, the center of Gaussian primitives are initialized at the locations of LiDAR points or 3D points obtained by running structure from motion.
The optimization of the parametrized $\mathcal{G}$ in 3D-GS is performed by projecting them into a set of camera planes associated with training images and subsequently minimizing image reconstruction error using differentiable rendering pipeline.

Some 3D reconstruction methods decompose the driving scene into a \emph{scene graph}.
This procedure gives a high degree of controllability, as each individual object is defined by a node in the graph~\cite{chen2024omnire, yan2024street, zhou2024hugs, zhou2024drivinggaussian}. 
We utilize OmniRe~\cite{chen2024omnire}, a recent driving scene reconstruction method which models the scene by a set of Gaussian nodes of different types.
Specifically, the static nodes $\mathcal{G}^s$ contribute to reconstruction of background, rigid nodes $\mathcal{G}^r$ contribute to reconstruction of foreground rigid objects such as cars, and morphable nodes $\mathcal{G}^m$ are used to model deformable objects such as pedestrians. 

In this formulation, the set of Gaussian primitives belonging to each set of rigid nodes $\mathcal{G}_i^r$ are due to a rigid (affine) transformation $\mathbf{T}_i=(\mathbf{R}_i,\mathbf{t}_i) \in \mathbb{SE}(3)$, so via a tensor product we have:
\begin{equation}
    \mathbf{T}_i \otimes \mathcal{G}_i = \{(o, \mathbf{R}_i\mathbf{\mu}+\mathbf{t}_i, Rot(\mathbf{R}_i,\mathbf{q}), \mathbf{s}, \mathbf{c})\},
\end{equation}
where $Rot(.)$ denotes rotating the quaternion by rotation matrix. 
The transformation $\mathbf{T}_i$ displaces all Gaussian points belonging to the $i^{th}$ object instance at each timestep, and is typically obtained by accessing the tracking information of 3D objects in the scene. 

\subsection{Gaussian-scene augmentation}
With access to a set of 3D Gaussian primitives, we can now ``splat'' them into the camera plane to render an image.
Let $\Phi$ denote the procedure of alpha-blending aggregation of depth-sorted Gaussian primitives $\mathcal{G}$ and their 3D-to-2D projection onto the 2D plane of a camera with known parameters $\omega$~\cite{kerbl20233d}. 
Given the camera parameters $\mathbf{\omega}_i$ corresponding to the training image $I_{i}$, reconstructing the image through the rendering process is defined as: 
\begin{equation}
    \hat{I}_{i} = \phi(\mathcal{G}^s, \mathcal{G}^r, \mathcal{G}^m, \mathbf{\omega}_i),
\end{equation}

Scene editing can be achieved by manipulating the Gaussian nodes in different ways. 
A common application is to modify camera parameters and synthesize novel views, but one
can also edit the scene by editing the graph nodes belonging to foreground objects $\{\mathcal{G}^r, \mathcal{G}^m\}$.
For example, the geometric transformation for the dynamic nodes can be modified ($\mathbf{T}_i+\Delta\mathbf{T}_i$) to edit object pose and position, one can delete objects from the scene by dropping their nodes in the graph ($\mathcal{G}^r - \{\mathcal{G}^r_k\}$), or one can add new nodes to insert a new object ($\mathcal{G}^r \cup \{\mathcal{G}^{a}\}$).
After modifying the set of primitives, rendering through $\Phi$ creates the frame and thus acts like a generator, allowing synthesis of new frames.

In this work, we focus on scene editing by object insertion, as represented in \cref{fig:appendix:overview_2}.
We add one or more new rigid nodes (called \emph{agents}) to the scene graph.
Given an agent object, represented by the set of Gaussian primitives $\mathcal{G}^{a}$ in its canonical coordinate system, object placement involves finding a physically plausible location for $\mathcal{G}^{a}$. 
Equivalently, this requires finding a rigid transformation $\mathcal{T}_{a} \in \mathbb{SE}(3)$ from a family of physically plausible rotation-translations.
Hence, rendering an augmented image $\hat{I}^{+}_{i}$, corresponding to $i^{th}$ camera, can be defined as follows:
\begin{equation} \label{eq:aug}
\begin{split}
 &\hat{I}^{+}_{i} = \phi(\mathcal{G}^s, \mathcal{G}^r \cup \mathcal{G}^{a} , \mathcal{G}^m, \mathbf{\omega}_i),\\
    &\mathcal{T}_{a}\otimes \mathcal{G}^{a} \rightarrow \mathcal{G}^{a}.
\end{split}
\end{equation}

For driving video augmentation, objects are typically not free to be placed anywhere. 
We want to place the object in an area that we refer to as \emph{drivable space}: visible in the camera plane, in a physically plausible location on the road, and not colliding with existing foreground objects or previously inserted agents.
This information can be obtained from the BEV road map and 3D bounding boxes of existing objects.
We simplify this problem by only allowing agent rotations around the gravity vector and inferring a plausible elevation from the annotations of surrounding objects, enabling us to parametrize $\mathcal{T}_{a}$ with only three elements. 
Sampling $\mathcal{T}_{a}$ is then possible by taking points uniformly at random from the drivable space.
If we find that the agent collides with existing objects, we reject the sample (\emph{collision} check).

It is still possible that the agent is placed correctly but is completely occluded by existing objects.
We therefore apply a visibility criterion ($\mathcal{V}_i$) that compares the depth image of the scene, rendered at the camera location, and the distance of Gaussian points in $\mathcal{G}_{a}$ to the camera. 
We define this as follows:
\begin{equation*}
r=\frac{\Sigma_i(\mathcal{V}_i)}{|(\mathcal{V}_i)|}, \textit{ where } \mathcal{V}_i= 
\begin{cases}
    1,& \text{if } z_i < D(u,v)\\
    0,              & \text{otherwise}
\end{cases}
\end{equation*}
Here, $z_i$ denotes the distance to camera of the $i^{th}$ point of the agent with center $\mathbf{u}_i=[x_i,y_i,z_i]$, and $D(u,v)$ denotes the rendered depth of the scene at pixel $(u,v)$ corresponding to the projection of $\mathbf{\mu}$ into the camera plane. 
If the visibility ratio ($r$) is smaller than a predefined threshold, it indicates the agent has been significantly occluded either by other foreground objects/vehicles or the structures in the scene, and we can again reject the sample (\emph{visibility} check).

\paragraph{Multiple cameras and agents.} All operations are directly applied on the 3D Gaussian field, which means that extending the setup to multi-camera image editing is trivial:
the rendering process of \cref{eq:aug} is a function of camera parameters ($\mathbf{\omega}$).
This is a major advantage of 3D modeling over implicit image generation methods that require cross-view consistent architectures~\cite{gao2023magicdrive}.

Inserting multiple agents into the scene is straightforward as well.
Each new agent is an additional node introduced to the scene graph. 
Of course, one should update the BEV map after placing each agent to avoid object collisions.

\section{Experiments}\label{sec:exp}

\subsection{Experimental Setup}

\paragraph{Dataset, 3D detectors, and metrics.}
We conduct experiments on nuScenes~\cite{caesar2020nuscenes}, consisting of 700 training and 150 validation scenes of resolution $900 \times 1600$.
Each scene recording comprises about 20 seconds of driving video, captured with 6 cameras at a frame rate of 10 HZ. 
The annotations are provided at frame rate of 2 Hz, in the form of 3D bounding boxes from 10 object categories. 

We report augmentation performance of our and competing methods of 3D object detectors, trained on augmented training frames and report their performance on the validation set of only real data.
We choose recent state-of-the-art 3D detectors in our evaluation, more specifically, FCOS3D~\cite{wang2021fcos3d} for monocular 3D detection and SparseBEV~\cite{liu2023sparsebev} for multi-camera 3D detection. 
We train both detectors using mmdetection3d~\cite{mmdet3d2020} with the basic configurations, without two-stage training or test-time augmentation.
Following common practice, we evaluate detector performance using the Mean Average Precision (mAP) and standard True Positive metrics such as Average Translation Error (ATE), Average Scale Error (ASE) and Average Orientation Error (AOE). 
We refer the reader to~\cite{caesar2020nuscenes} for details about these metrics.

\begin{table*}[t]
\caption{\small The results of 3D data augmentation on  monocular 3D object detection on nuScenes validation set.}
  \label{tab:results_fcos3d}
  \vspace{-8pt}
  \centering
\resizebox{\linewidth}{!}{
  \begin{tabular}{ll|cccc|cccccccccc}
    \toprule
     &\multirow{2}{*}{Data} & \multirow{2}{*}{mAP$\uparrow$} & \multirow{2}{*}{mATE$\downarrow$} & \multirow{2}{*}{mASE$\downarrow$} & \multirow{2}{*}{mAOE$\downarrow$} & \multicolumn{10}{c}{APs $\uparrow$} \\
    & & & & & & car & truck & bus & trailer & constr. & pedest. & motorc. & bicycle & tr. cone & barrier \\
    \midrule
    \midrule
    & Real data (1x) & 31.16 & 0.7691 & 0.2643 & 0.4848 & 48.0 & 25.8 & 29.9 & 10.0 & 5.5 & 40.5 & 29.9 & 26.2 & 52.3 & 43.6 \\
    \cmidrule{2-16}
    \multirow{4}{*}{\rotatebox[origin=c]{90}{Augm. only}}
    & MagicDrive~\cite{gao2023magicdrive} (50\% boxes)  & 10.77 & 1.0278 & 0.3705 & 0.8002 & 25.4 & 6.5 & 7.4 & 0.4 & 0.8 & 17.4 & 9.4 & 6.2 & 11.4 & 22.8 \\
    & MagicDrive~\cite{gao2023magicdrive} (100\% boxes) & 13.77 & 0.9682 & 0.3225 & 0.7156 & 29.3 & 8.5 & 9.0 & 2.2 & 1.4 & 21.5 & 12.4 & 7.7 & 20.2 & 25.4 \\
    & Depth-Cond. Inpainting                            & 26.25 & 0.8393 & 0.2717 & 0.5541 & 44.7 & 18.4 & 19.6 & 5.3 & 3.1 & 39.1 & 25.5 & 21.9 & 50.6 & 34.1 \\
     \cmidrule{2-16}
    & 3D-GS & \textbf{31.48} & 0.7809 & 0.2663 & 0.4126 & 47.1 & 25.5 & 35.8 & 9.6 & 6.9 & 40.0 & 29.7 & 26.1 & 51.4 & 42.7 \\
    \midrule
    \midrule
    & Real data (2x) & 32.57 & 0.7543 & 0.2599 & 0.4287 & 49.3 & 27.1 & 34.9 & 10.5 & 6.1 & 40.9 & 30.2 & 28.0 & 53.1 & 45.6 \\
     \cmidrule{2-16}
    \multirow{4}{*}{\rotatebox[origin=c]{90}{Real + augm.}}
    & Real + MagicDrive~\cite{gao2023magicdrive} (50\% boxes) & 30.42 & 0.7765 & 0.2586 & 0.4788 & 47.7 & 25.5 & 31.3 & 8.7 & 5.4 & 39.3 & 29.4 & 23.0 & 50.7 & 43.1 \\
    & Real + MagicDrive~\cite{gao2023magicdrive} (100\% boxes) & 30.52 & 0.7607 & 0.2572 & 0.5206 & 46.8 & 25.7 & 32.2 & 12.1 & 5.7 & 38.0 & 28.8 & 23.1 & 49.0 & 43.8 \\
    & Real + Depth-Cond. Inpainting & 31.66 & 0.7550 & 0.2649 & 0.4368 & 48.5 & 26.3 & 31.2 & 10.7 & 5.6 & 40.9 & 30.9 & 26.0 & 52.1 & 44.3 \\
    \cmidrule{2-16}
    & Real + 3D-GS & \textbf{33.20} & 0.7395 & 0.2590 & 0.3817 & 49.2 & 28.2 & 34.7 & 12.6 & 5.7 & 41.4 & 30.9 & 29.6 & 54.3 & 45.5 \\
    \bottomrule
  \end{tabular}}
\end{table*}
\begin{figure*}[t!]
\resizebox{\linewidth}{!}{
\begin{tabular}{llll}
\includegraphics[width=.3\linewidth]{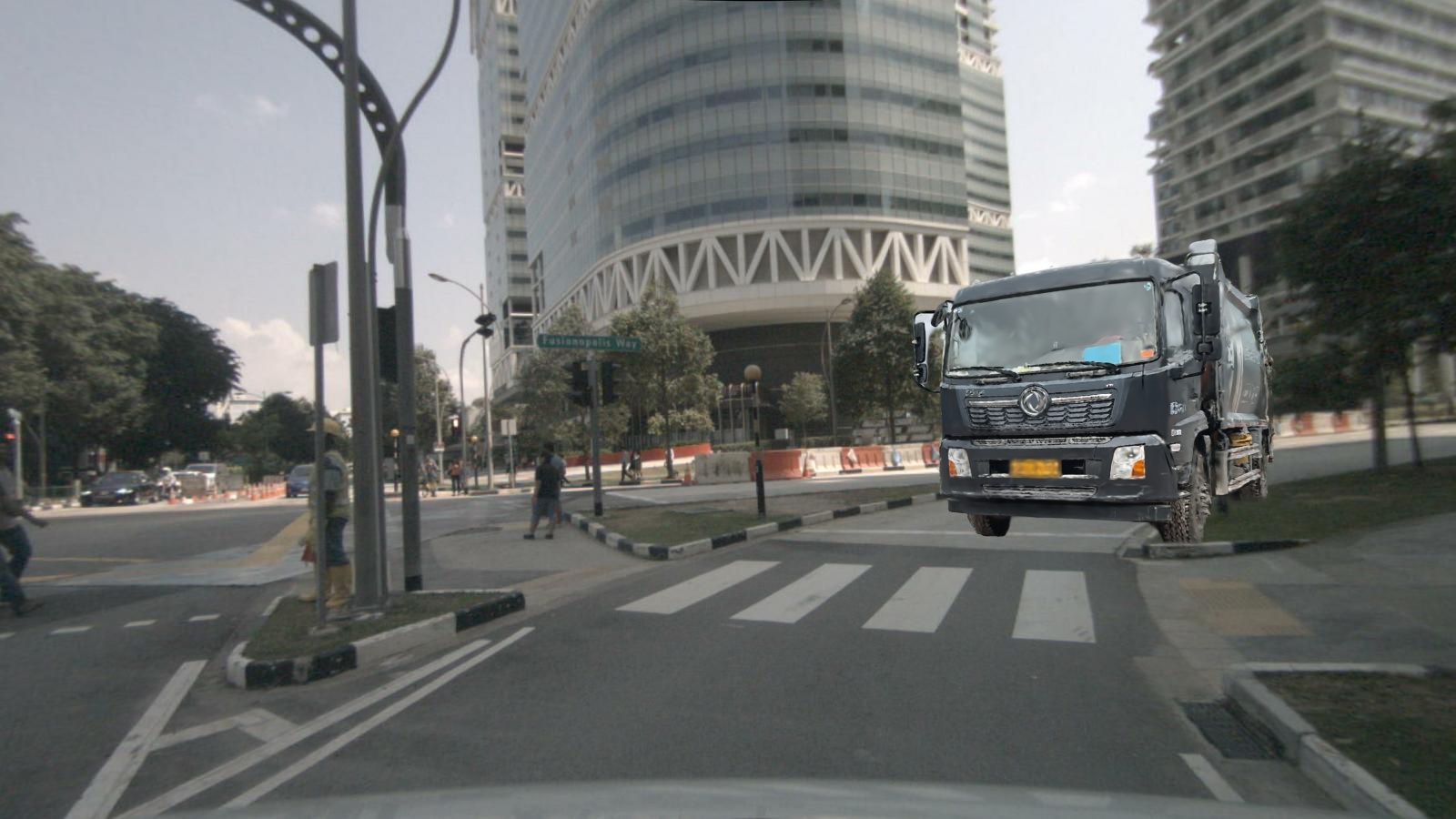}&
\includegraphics[width=.3\linewidth]{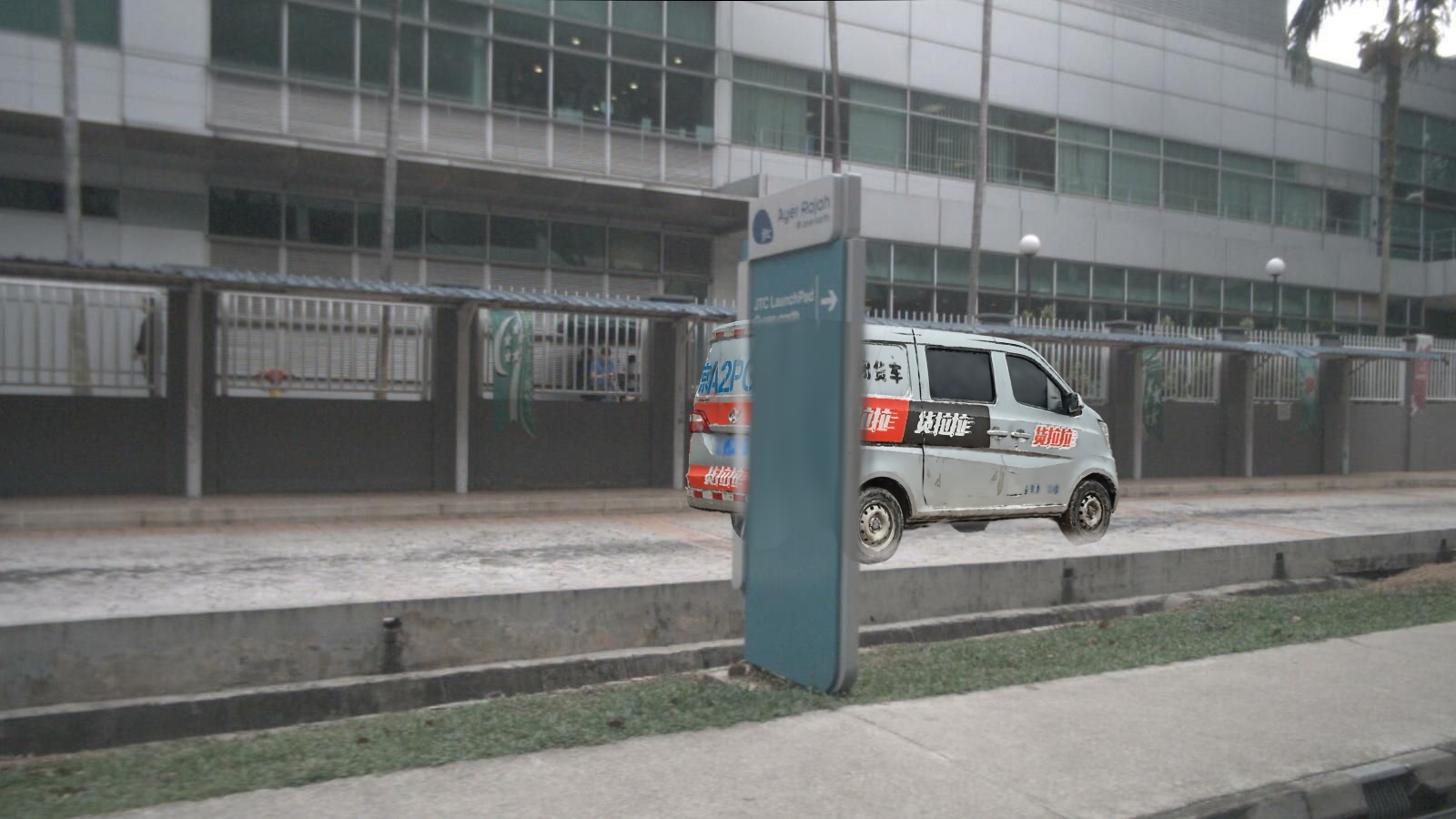}&
\includegraphics[width=.3\linewidth]{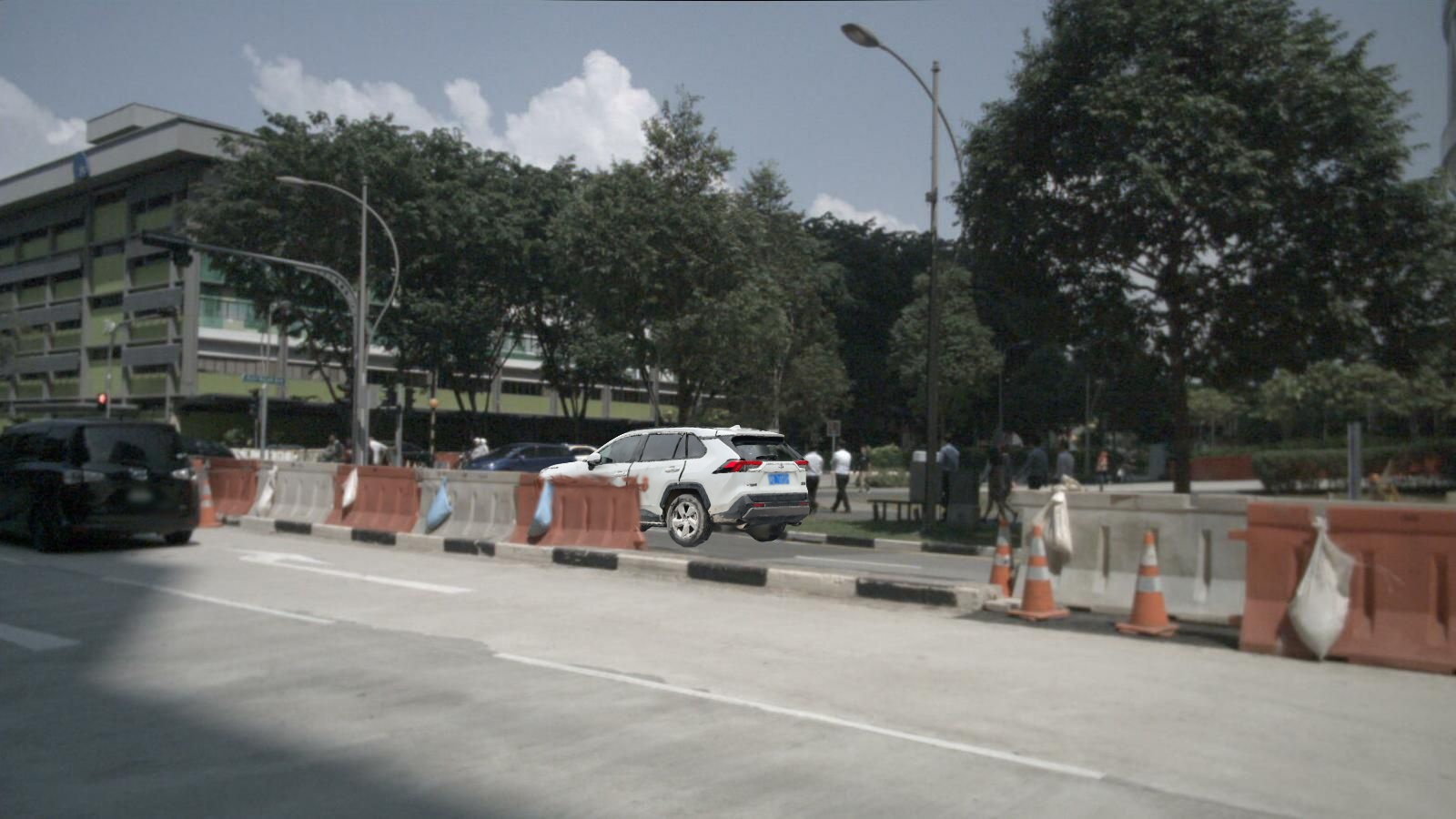}&
\includegraphics[width=.3\linewidth]{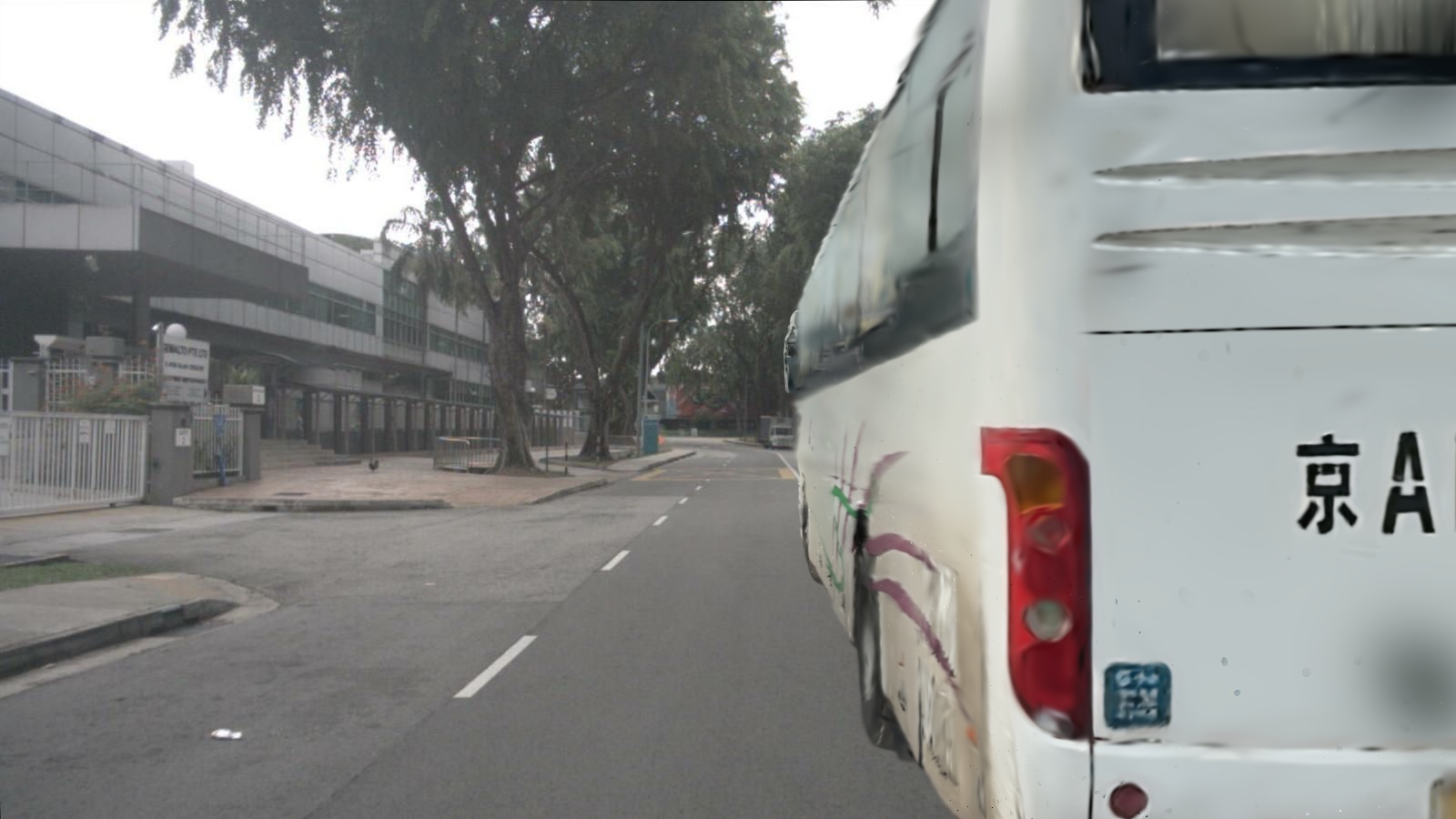}
\end{tabular}
}
\vspace{-10pt}
\caption{\small Examples of single-camera augmentation. This examples demonstrate physically plausible insertion of one agent in camera view.}
\vspace{-10pt}
   \label{fig:search_vs_optaug}
\end{figure*}
\paragraph{3D scene modeling and synthetic agents library.}
For 3D modeling of street scenes, we use OmniRe~\cite{chen2024omnire}, a reconstruction framework based on instance-aware 3D Gaussian Splatting.
This framework enables real-time ($\sim$60 Hz) rendering of scenarios.
To obtain synthetic 3D agents of all object categories for augmentation, we use various multi-view image datasets and a standard 3D-GS pipeline to create a set of fully-reconstruction objects. 
For car, bus and truck classes, we use images from the 3DRealCars~\cite{du20243drealcar} dataset. 
For bicycles and motorcycles, we use CO3D(v2)~\cite{reizenstein21co3d}.
For trailer, barrier, construction vehicle, and traffic cone classes, since there is no public multi-view image datasets, we utilize publicly available 3D meshes,\footnote{Meshes from https://sketchfab.com} convert them into multi-view images using rendering engines, and perform 3D reconstruction using the 3D-GS pipeline (more details are given in the Appendix). 
In total, we obtain around 10 objects per object category, excluding pedestrians, for which we do not add synthetic agents.
Although integration of different sources was required to build the asset library, we will show that even just a limited number of instances is sufficient for data augmentation purposes. 

\paragraph{Baselines.}
We use two recent diffusion-based methods as baselines, namely MagicDrive~\cite{gao2023magicdrive} and 
its follow-up video model, MagicDriveDiT~\cite{gao2024magicdrivedit}, both of which can produce multi-view generations. 
Both approaches generate frames conditioned on the birds eye view map and a given set of 3D bounding boxes. 
We utilize publicly released checkpoints, generating frames with 272$\times$736 and 424$\times$800 resolution, respectively.
Following the practice suggested by MagicDrive authors, we condition synthesis on training layouts after dropping 50\% of the 3D boxes, and resize and pad the frames to $900 \times 1600$ resolution before training.
For completeness, we also include an experiment where the layout contains 100\% of 3D boxes. 
We further include a baseline where we replace all labeled objects in training frames with a pretrained Stable Diffusion 2 inpainting model~\cite{rombach2022stablediffusion}, conditioned on depth maps through a ControlNet~\cite{controlnet}.
This procedure has the effect of augmenting the appearance of street objects, leaving the layout of the scene and its background intact.
As there is no trivial way to guarantee consistency in appearance of the same object across multiple views, we test this baseline solution only in the monocular data augmentation and detection setting.

\subsection{Results}
\begin{table*}[t]
\footnotesize
\caption{\small The results of 3D data augmentation on  multi-camera 3D object detection on nuScenes validation set.}
\vspace{-8pt}
  \label{tab:results_sparsebev}
  \centering
  \begin{adjustbox}{width=\linewidth}
  \begin{tabular}{ll|cccc| M{6mm} M{6mm} M{6mm} M{6mm} M{9mm} M{9mm} M{9mm} M{9mm} M{8mm} M{8mm}}
    \toprule
   &\multirow{2}{*}{Data} & \multirow{2}{*}{mAP$\uparrow$} & \multirow{2}{*}{mATE$\downarrow$} & \multirow{2}{*}{mASE$\downarrow$} & \multirow{2}{*}{mAOE$\downarrow$} & \multicolumn{10}{c}{APs $\uparrow$} \\
    & & & & & & car & truck & bus & trailer & constr. & pedest. & motorc. & bicycle & tr.cone & barrier \\
    \midrule
    \midrule
     & Real data (1x)                           & 33.93 & 0.7312 & 0.2748 & 0.5549 & 53.4 & 25.2 & 33.9 & 8.6 & 5.0 & 42.4 & 34.2 & 27.8 & 58.8 & 50.1 \\
    \cmidrule{2-16}      
    \multirow{5}{*}{\rotatebox[origin=c]{90}{Augm. only}}
     & MagicDrive~\cite{gao2023magicdrive} (50\% boxes)              & 24.70 & 0.7985 & 0.2769 & 0.7168 & 43.3 &  18.1 & 24.6 & 4.0 & 2.5 & 29.2 & 19.4 & 19.1 & 45.4 & 41.4 \\
     & MagicDrive~\cite{gao2023magicdrive} (100\% boxes)             & 26.69 & 0.7807 & 0.2788 & 0.7003 & 46.5 & 19.7 &  24.2 & 5.8 & 2.6 & 29.4 & 23.9 &  21.8 & 48.3 & 44.9 \\
     & MagicDriveDiT~\cite{gao2024magicdrivedit} (50\% boxes)           & 20.95 & 0.9039 & 0.7105 & 1.5436 & 39.1 & 15.2 & 17.5 & 2.7 & 1.3 & 25.5 & 15.2 & 14.7 & 39.5 & 39.0 \\
     & MagicDriveDiT~\cite{gao2024magicdrivedit} (100\% boxes)          & 21.89 & 0.8722 & 0.2850 & 0.7167 & 41.7 & 16.6 & 22.5 & 3.7 & 1.9 & 26.1 & 13.8 & 12.9 & 40.1 & 39.6 \\
     \cmidrule{2-16}                          
     & 3D-GS              & \textbf{34.97} & 0.7122 & 0.2708 & 0.4880 & 52.9 & 28.5 & 36.3 & 9.5 & 4.2 & 42.7 & 32.2 & 30.0 & 58.7 & 54.7 \\
    \midrule                                  
    \midrule                                  
     & Real data (2x)                    & 35.77 & 0.6910 & 0.2736 & 0.5116 & 56.0 & 27.6 & 38.8 & 13.6 & 5.7 & 42.4 & 35.3 & 26.9 & 59.7 & 51.8 \\
    \cmidrule{2-16}
    \multirow{5}{*}{\rotatebox[origin=c]{90}{Real + augm.}}
     & Real + MagicDrive~\cite{gao2023magicdrive} (50\% boxes)      & 34.19 & 0.7039 & 0.2722 & 0.5246 & 53.8 & 25.6 & 38.1 & 9.4 & 5.4 & 41.2 & 33.7 & 25.8 & 57.8 & 51.2 \\
     & Real + MagicDrive~\cite{gao2023magicdrive} (100\% boxes)     & 32.73 & 0.7011 & 0.2764 & 0.5427 & 53.6 & 24.0 & 35.9 & 10.8 & 4.7 & 37.6 & 30.2 & 25.6 & 56.4 & 48.4 \\
     & Real + MagicDriveDiT~\cite{gao2024magicdrivedit} (50\% boxes)   & 34.07 & 0.7279 & 0.2766 & 0.5816 & 54.0 & 26.0 & 33.2 & 10.3 & 4.8 & 42.3 & 32.9 & 25.6 & 58.5 & 53.0 \\
     & Real + MagicDriveDiT~\cite{gao2024magicdrivedit} (100\% boxes)  & 31.48 & 0.7001 & 0.2766 & 0.6465 & 53.4 & 23.2 & 31.0 & 9.9 & 5.0 & 38.0 & 29.5 & 21.9 & 54.6 & 48.2 \\
     \cmidrule{2-16}
     & Real + 3D-GS      & \textbf{37.17} & 0.6825 & 0.2715 & 0.4460 & 56.1 & 29.6 & 42.6 & 13.9 & 5.7 & 43.3 & 35.7 & 30.8 & 60.8 & 53.3 \\
    \bottomrule
  \end{tabular}
  \end{adjustbox}
  
\end{table*}
\cref{tab:results_fcos3d} and \cref{tab:results_sparsebev} present the augmentation performance in monocular and multi-camera 3D detections, respectively.
In both settings, we experiment with two augmentation schemes, i.e., using only augmented frames (top section) and using a mixture of clean real frames and augmented ones (bottom section).
We produce one augmented frame for every training frame, which means the mixture setting (``Real + augm.'') has twice the amount of training data.
In this setting we also train for twice as many update steps.

From the monocular 3D object detection results \cref{tab:results_fcos3d}, we observe that augmentation by Gaussian Splatting proves more beneficial than diffusion-based methods. 
In fact, our augmentation approach is the only solution outperforming the baseline of only using real data, both improving mAP and reducing average errors on nuScenes true positive metrics. 
We ascribe the limited performance of MagicDrive in augmentation to the discrepancies in resolution between its generations and the nuScenes validation set, as well as the occasional limited geometric fidelity to the desired layout.
For the depth conditioned inpainting baseline, we believe its limited augmentation performances are due to the fact that by construction it can vary the objects in appearance, but not in 3D geometry.
More broadly, \cref{tab:results_fcos3d} suggests that training on a mixture of real and augmented data is a much more successful scheme than to training on augmented data only.
Some examples of synthetic agents added into monocular camera views are represented in \cref{fig:search_vs_optaug}.

\begin{figure*}[t]
\begin{center}
  \includegraphics[width=0.95\linewidth, trim={0cm, 4cm, 0cm, 4cm}, clip]{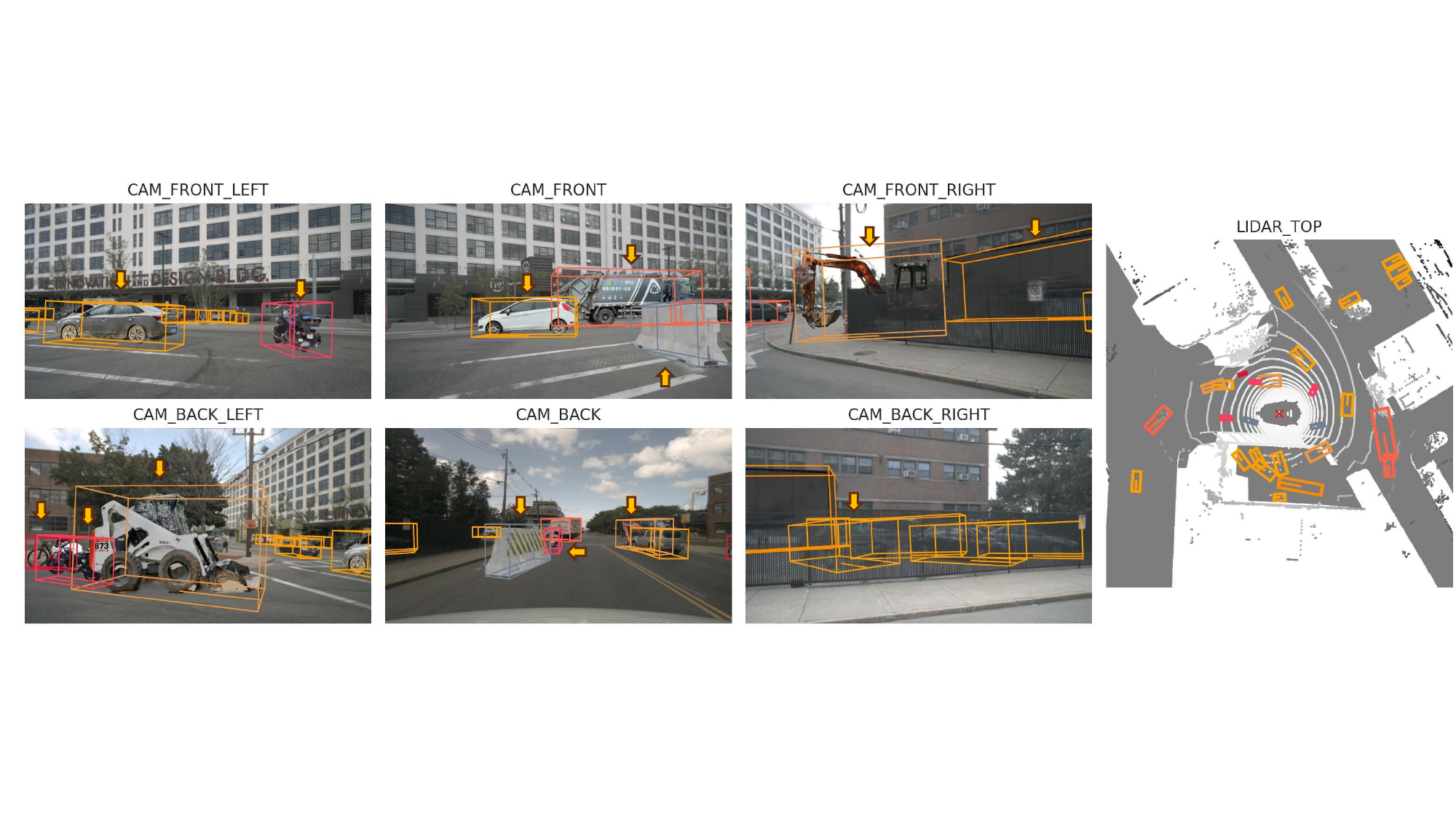}\\
  \includegraphics[width=0.95\linewidth, trim={0cm, 4cm, 0cm, 4cm}, clip]{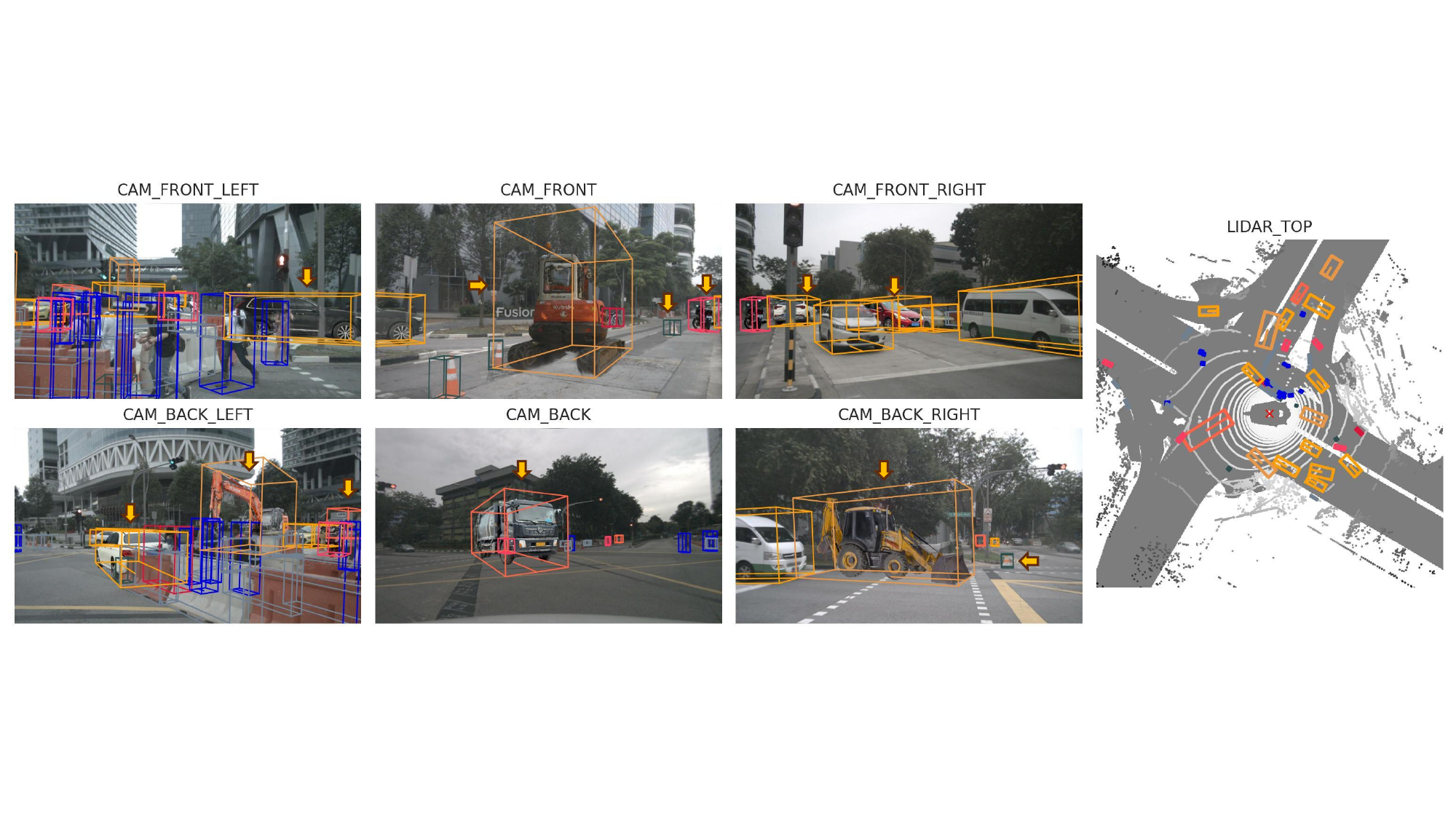}\\
  \includegraphics[width=0.95\linewidth, trim={0cm, 4cm, 0cm, 4cm}, clip]{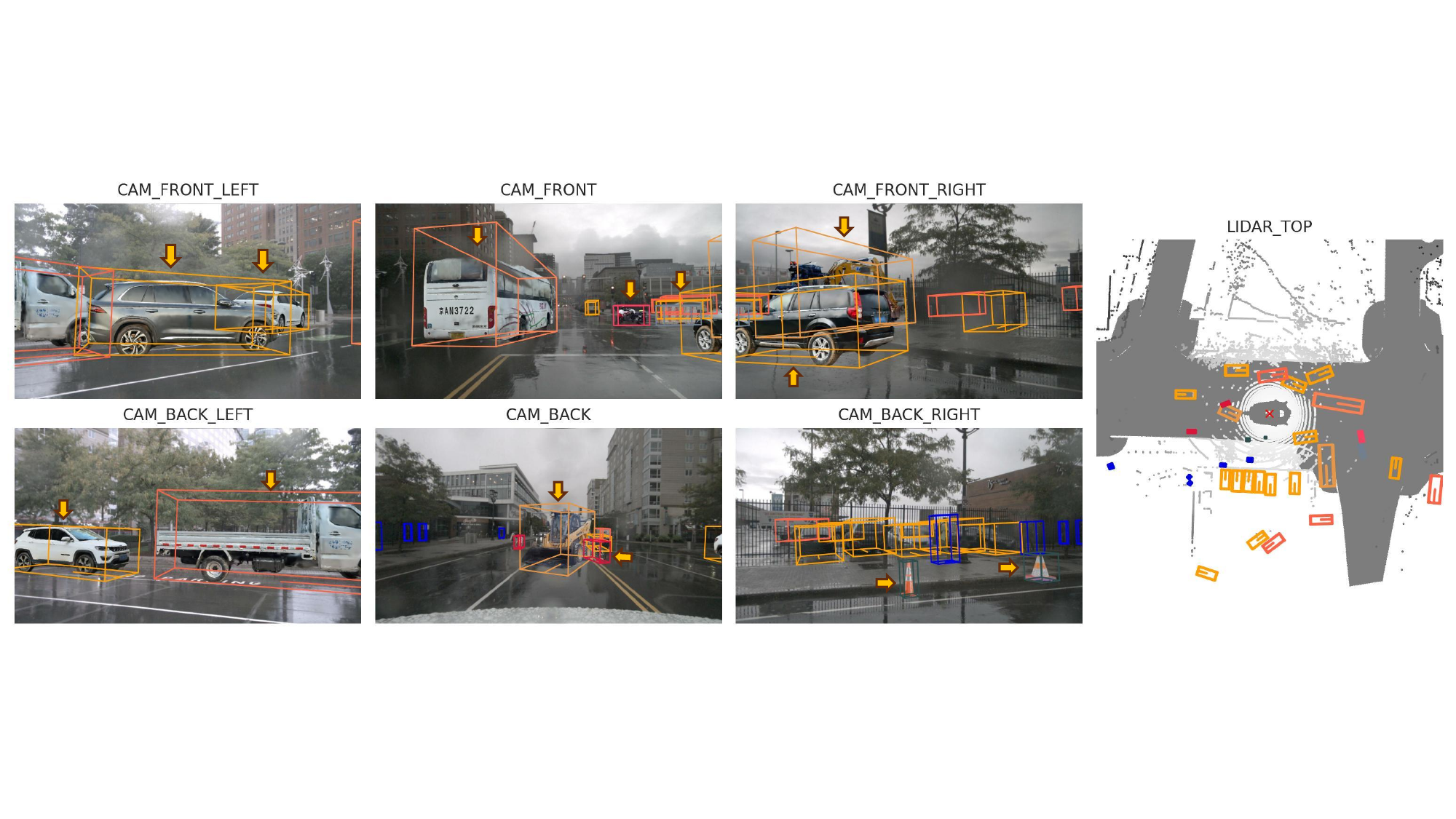}
\end{center}
\vspace{-10pt}
\caption{\small Examples of multi-camera 3D augmentation and their associated  3D bounding boxes. The arrows indicate the inserted agents.}
   \label{fig:qualitative_augmentation}
\vspace{-5pt}
\end{figure*}

Similar observations can be made by looking at \cref{tab:results_sparsebev}, showcasing the performance on augmenting a SparseBEV model on multi-view object detection.
Again, using augmented data in a mixture with real frames proves to be a more successful strategy than using augmented data only.
Consistently with the monocular detection benchmark, augmentation by Gaussian Splatting outperforms diffusion-based strategies.
Interestingly, MagicDrive outperforms its own follow-up work MagicDriveDiT. 
We suspect that this counterintuitive result comes from MagicDriveDiT being trained as a video model, therefore exhibiting slighter worse quality for individual frames.
\cref{fig:qualitative_augmentation} shows some multi-camera frames augmented by 3D-GS.

\subsection{Geometric vs. photometric diversity}
\begin{table}[t]
\caption{Impact of number of synthetic 3D agents over augmentation performances. Number of agents is reported as \# added agents per camera / \# of unique agents per class.}
\label{tab:ablation_agents}
\centering
\resizebox{\linewidth}{!}{
\begin{tabular}{l|ccc|cccc}
\toprule
&\multirow{2}{*}{real} & \multirow{2}{*}{augm.} & num.& \multirow{2}{*}{mAP$\uparrow$} & \multirow{2}{*}{mATE$\downarrow$} & \multirow{2}{*}{mASE$\downarrow$} & \multirow{2}{*}{mAOE$\downarrow$}\\
&&& agents \\
\midrule
\rowcolor{Gray}\cellcolor{white}
\multirow{3}{*}{FCOS3D~\cite{wang2021fcos3d}} 
& \cmark & \xmark & - & 32.57 & 0.7543 & 0.2599 & 0.4287\\
&\multirow{2}{*}{\cmark}&\multirow{2}{*}{\cmark}& 1/1 & 32.96 & 0.7573 & 0.2577 &  0.4457 \\
&&& 3/10 & 33.20 & 0.7395 & 0.2590 & 0.3817 \\
\midrule
\rowcolor{Gray}\cellcolor{white}
\multirow{3}{*}{SparseBEV~\cite{liu2023sparsebev}} 
& \cmark & \xmark & - & 35.77 & 0.6910 & 0.2736 & 0.5116\\
&\multirow{2}{*}{\cmark}&\multirow{2}{*}{\cmark}& 1/1 & 36.94 & 0.6805 & 0.2719 & 0.4453 \\
&&& 3/10 & 37.17 & 0.6825 & 0.2715 & 0.4460 \\
\bottomrule
\end{tabular}}
\end{table}
In our experiments in \cref{tab:results_fcos3d,tab:results_sparsebev}, diffusion-based methods such as depth-conditioning inpainting, MagicDrive, and MagicDriveDiT introduce high photometric diversity to the dataset. 
In contrast, the proposed 3D-GS approach accessed only a limited number of 3D reconstructed objects, approximately 10 agents per object category. 
This raises the question of whether the superior performance of Gaussian Splatting is due to the application of diverse geometric transformations to the inserted objects.

To investigate this issue, we conducted an experiment with an extreme setup by incorporating only a unique object per object category to augment the entire dataset.
The results of this setup, which implies minimal photometric diversity in our augmentation pipeline, are reported in \cref{tab:ablation_agents}. 
Our experiments reveal two interesting insights: (1) even with severely bounded photometric diversity, the Gaussian Splatting augmentation outperforms using only real data as well as all diffusion-based approaches in \cref{tab:results_fcos3d,tab:results_sparsebev} by a significant margin, and (2) when moving from augmenting with a unique agent to employing all the agents, the performance gain of the detector is rather small, indicating that increasing the visual diversity of assets helps, but only marginally.

\begin{table*}[t]
\caption{\small The impact of applied low/high pose variation and low/high occlusion for multi-camera 3D object detection. All 3D augmentation experiments are performed by placing one agent per camera view. The obtained results on the Nuscene validation set are shown. }
\vspace{-8pt}
  \label{tab:ablation}
  \centering
  \begin{adjustbox}{width=\linewidth}
  \begin{tabular}{ll|cccc|cccccccccc}
    \toprule
    \multirow{2}{*}{} & \multirow{2}{*}{Data} & \multirow{2}{*}{mAP$\uparrow$} & \multirow{2}{*}{mATE$\downarrow$} & \multirow{2}{*}{mASE$\downarrow$} & \multirow{2}{*}{mAOE$\downarrow$} & \multicolumn{10}{c}{APs $\uparrow$} \\
    & & & & & & car & truck & bus & trailer & constr. & pedes. & motorc. & bicycle & tr. cone & barrier \\
    \midrule
      & Real data (2x)  & 35.77 & 0.6910 & 0.2736 & 0.5116 & 56.0 & 27.6 & 38.8 & 13.6 & 5.7 & 42.4 & 35.3 & 26.9 & 59.7 & 51.8 \\
    \cmidrule{2-16}
     \multirow{4}{*}{\rotatebox[origin=c]{90}{Real+augm.}} & Real + pose-aligned      & 36.44 & 0.6870 & 0.2689 & 0.4758 & 55.9 & 27.3 & 38.8 & 14.1 & 5.9 & 43.3 & 35.6 & 30.5 & 59.7 & 53.4 \\
      & Real + pose-rotated      & 36.44 & 0.6847 & 0.2725 & \textbf{0.4389} & 55.7 & 27.7 & 40.2 & 14.8 & 5.5 & 42.9 & 35.8 & 28.0 & 59.4 & 54.4 \\
     \cmidrule{2-16}
      & Real + low occlusion     & 36.51 & 0.6892 & 0.2677 & 0.4605 & 55.7 & 27.8 & 40.0 & 13.8 & 5.5 & 43.1 & 35.0 & 29.8 & 60.1 & 54.3 \\
      & Real + high occlusion    & 36.41 & 0.6906 & 0.2713 & \textbf{0.4372} & 56.1 & 28.3 & 38.6 & 13.0 & 5.2 & 43.4 & 36.7 & 30.3 & 60.1 & 52.4 \\
    \bottomrule
  \end{tabular}
  \end{adjustbox}
  
\end{table*}

\begin{figure*}[!t]
\resizebox{\linewidth}{!}{
\begin{tabular}{ll}
\includegraphics[width=.5\linewidth, trim={0cm 0 10cm 0cm}, clip]{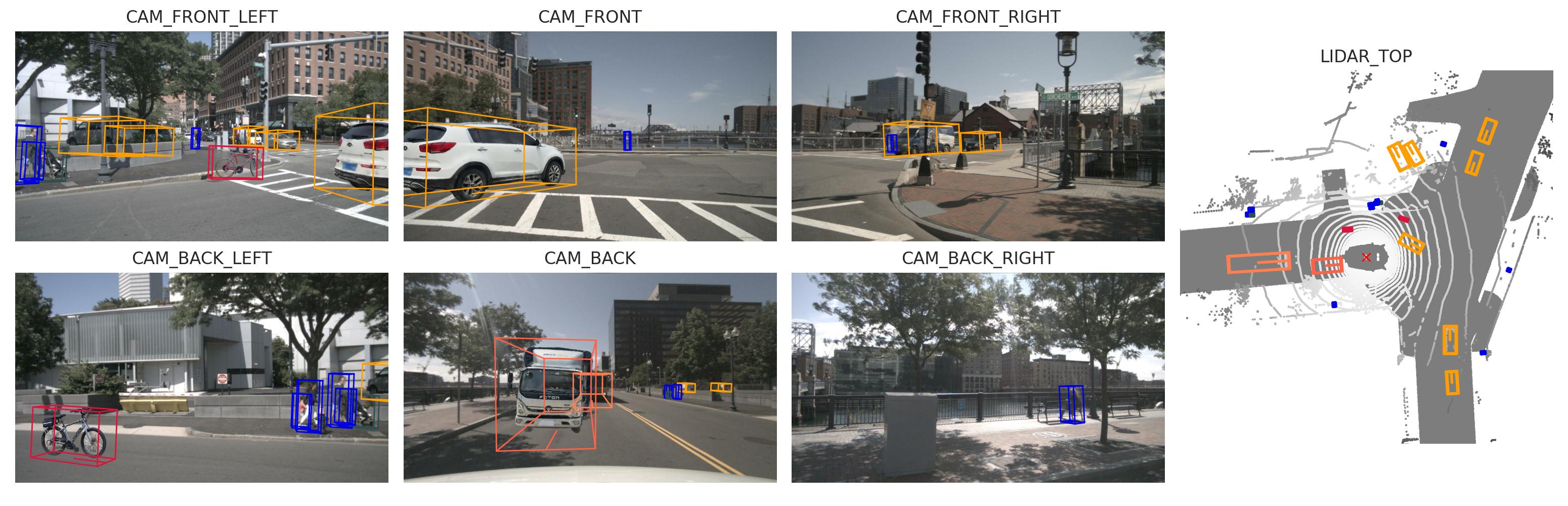}&
\includegraphics[width=.5\linewidth, trim={0cm 0 10cm 0cm}, clip]{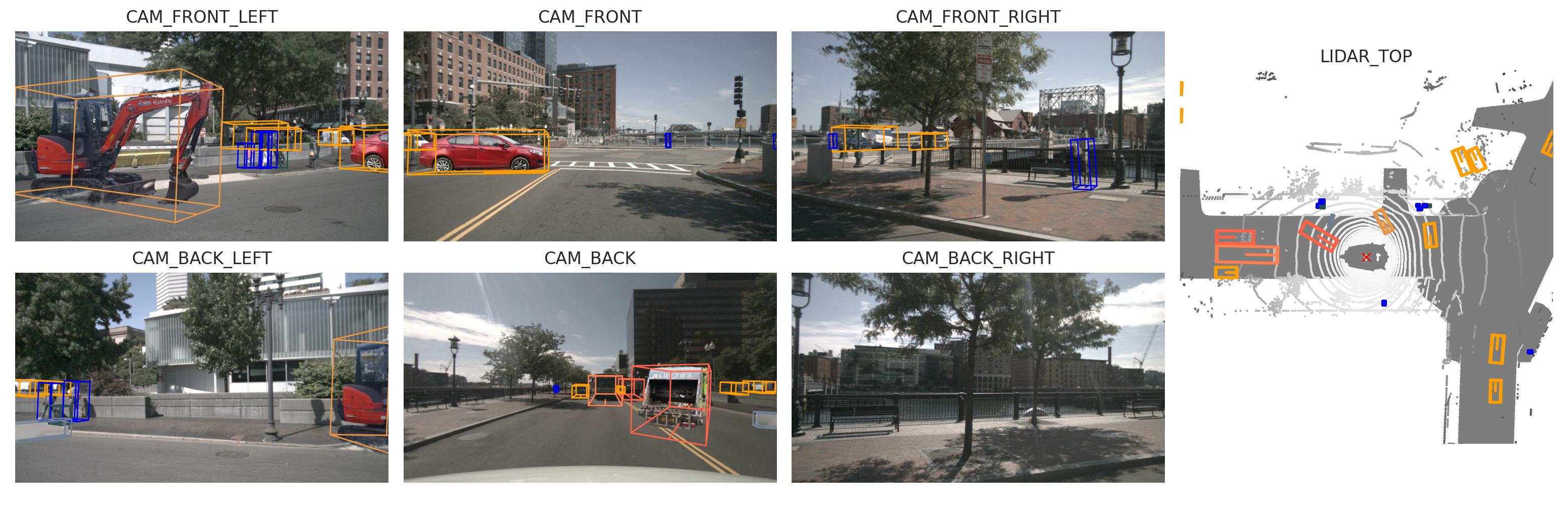}
\end{tabular}
}
\vspace{-15pt}
\caption{\small Examples of pose-aligned agent placement (left) versus random pose placement (right).}
   \label{fig:agent_pose_var}
\end{figure*}
\begin{figure*}[!t]
\resizebox{\linewidth}{!}{
\begin{tabular}{ll}
\includegraphics[trim={1cm 0 10cm 0cm}, clip, width=0.5\linewidth]{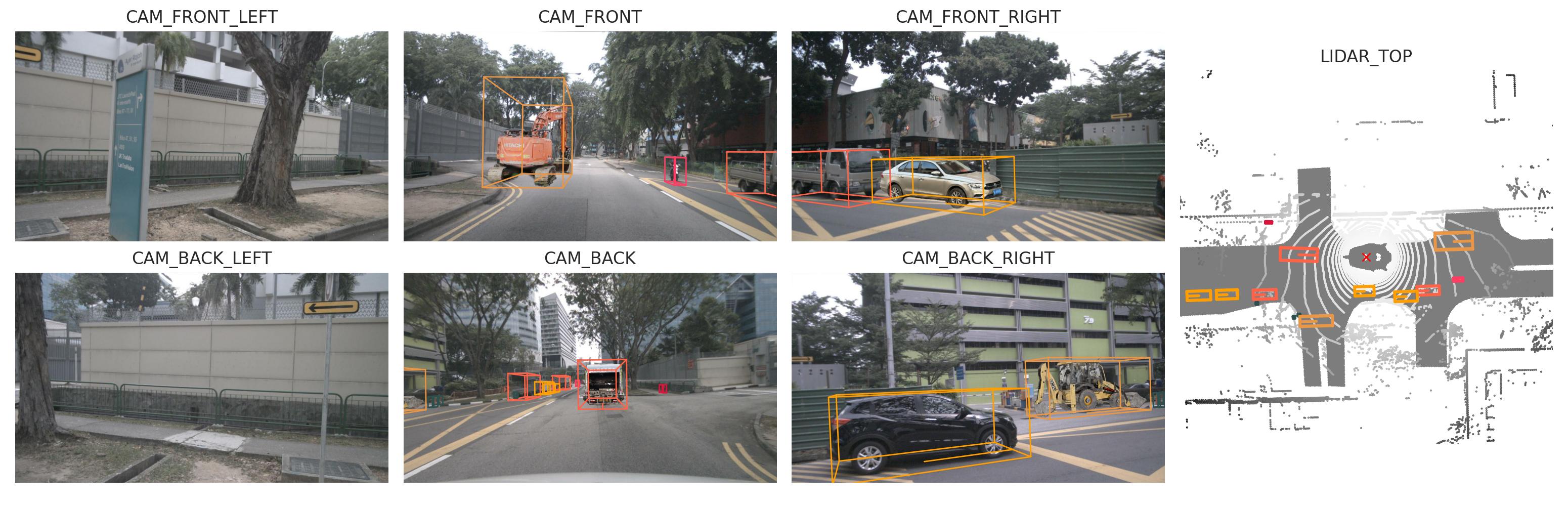}&
\includegraphics[trim={1cm 0 10cm 0cm}, clip, width=0.5\linewidth]{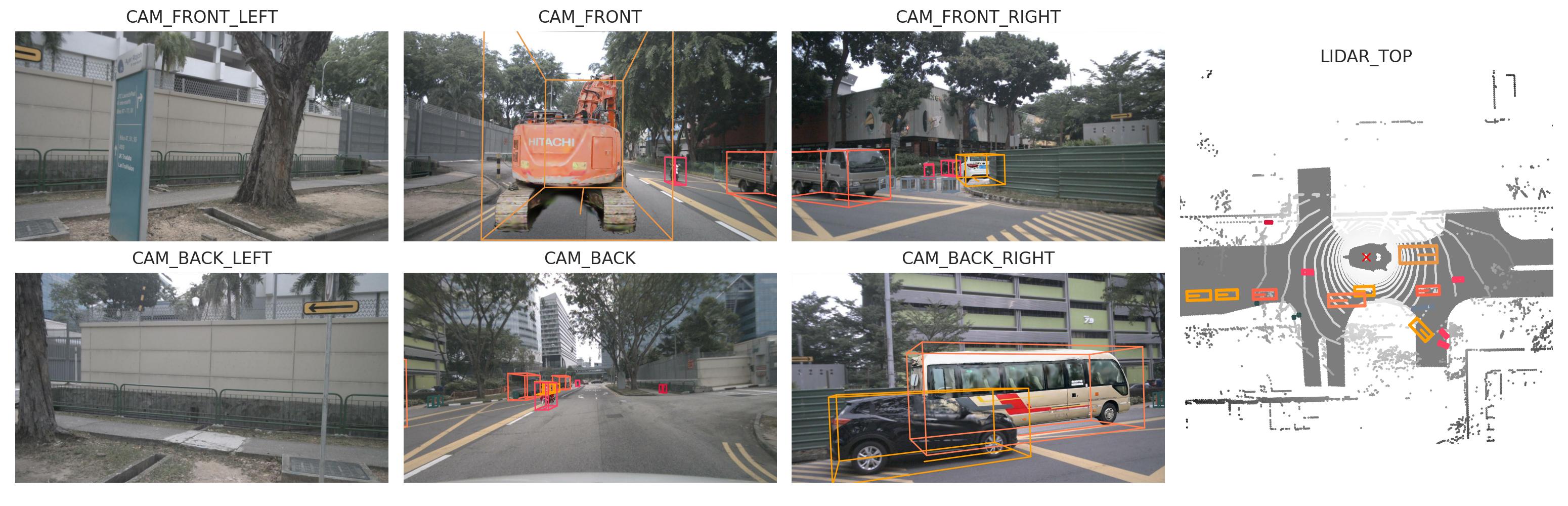}
\end{tabular}
}
\vspace{-15pt}
\caption{\small Examples of agent placement with low (left) and high (right) occlusion score.}
   \label{fig:occlusion}
\vspace{-10pt}   
\end{figure*}

\subsection{Augmenting with pose variation}
A control experiment is setup to investigate the impact of imposing minimum and maximum pose variations in the process of agent placement on the data augmentation. 
For imposing a minimum pose variation to the augmented objects, the pose of the closest object to the agent, based on their Euclidean distance in the scene, is being used for placement. Therefore, this introduces a minimal pose variation with respect to the pose of existing object in the camera view. Alternatively, assigning a random pose in the range of $(0, 2\pi)$ to the agent in every frame, impose a maximum pose variation in the augmentation process. \cref{tab:ablation} shows the detection performance after training on the mixture of real and augmented samples with these two different strategies. While the mAP does not show any difference, the AOE significantly improves when various agent poses are included in the augmented data. 
\cref{fig:agent_pose_var} visualizes two examples of aligned-pose versus random-pose placement of agents into the scene.

\subsection{Augmenting with hard examples}
\begin{table}[t]
\caption{Results on the impact of the location of synthetic agents. "Hard" locations maximize the loss of the respective detector.}
\label{tab:ablation_location}
\centering
\resizebox{\linewidth}{!}{
\begin{tabular}{l|cc|cccc}
\toprule
& \multirow{2}{*}{Setting} & Agent & \multirow{2}{*}{mAP$\uparrow$} & \multirow{2}{*}{mATE$\downarrow$} & \multirow{2}{*}{mASE$\downarrow$} & \multirow{2}{*}{mAOE$\downarrow$}\\
&& location \\
\midrule
\multirow{4}{*}{\rotatebox[origin=c]{90}{\small FCOS3D~\cite{wang2021fcos3d}}}
& \multirow{2}{*}{Augm. only} & random &  31.48 & 0.7809 & 0.2663 & 0.4126 \\
&& hard & 30.61 & 0.7601 & 0.2636 & 0.4400\\
\cmidrule{2-7}
& \multirow{2}{*}{Real + augm.} & random & 33.20 & 0.7395 & 0.2590 & 0.3817 \\
&& hard & 33.51 & 0.7576 & 0.2595 & 0.3954 \\
\midrule
\multirow{4}{*}{\rotatebox[origin=c]{90}{\small Sp. BEV~\cite{liu2023sparsebev}}}
& \multirow{2}{*}{Augm. only} & random & 34.97 & 0.7122 & 0.2708 & 0.4880  \\
&& hard & 34.57 & 0.7181 & 0.2752 & 0.5279 \\
\cmidrule{2-7}
& \multirow{2}{*}{Real + augm.} & random & 37.17 & 0.6825 & 0.2715 & 0.4460 \\
&& hard & 36.73 & 0.6820 & 0.2710 & 0.4434 \\
\bottomrule
\end{tabular}}
\vspace{-.5cm}
\end{table}

An alternative to random object placement in scenes is generating hard examples for dataset augmentation. 
While numerous studies have reported the positive impact of generating hard examples (e.g., through adversarial attacks or hard-mining methods) for augmentation in 2D perception tasks such as classification~\cite{athalye2018synthesizing, hemmat2023feedback} or 2D object detection~\cite{wang2024detdiffusion, zhou2023f}, 
there are only few works investigating this for 3D object detection~\cite{lehner20223d}, and to the best of our knowledge none for the camera-only detection setting.

To investigate the the usefulness of \emph{hard} samples on detection performance, we implement two alternative approaches to random agent placement. In the first approach, by leveraging the fast rendering pipeline of 3D-GS, the prediction error of the detector is used to identify the most challenging edited image for 3D detection. 
The exploration is performed on a set of 16 random seeds over drivable space, computing the loss value of the 3D detector on each corresponding generated image. The corresponding image with the highest loss value is considered a hard example and is later used for data augmentation. We can specify a hard example as follows:
\begin{equation}\label{eq:search}
    \operatorname*{argmax}_{(\hat{I}^{+},y^{+})} \left(\mathcal{L}(f_\psi(\hat{I}^{+}), y^{+})\right),
\end{equation}
where $y^{+}$ denotes updated 3D object annotations after placing the agent into the scene and $f_\psi$ is the 3D detector.

\cref{tab:ablation_location} shows augmenting the dataset with hard examples, guided by detection loss, doesn't consistently outperform random placement in both single-camera and multi-camera cases. 
Our speculation on the superior performance of random placement is due to the inclusion of both easy and hard samples, which are more easily learnable after augmentation.
Differently, when generating only hard examples in the augmentation process mostly challenging cases are included, which the 3D detectors can't leverage fully to enhance its performance. 

In the second approach, we define the hard examples based on the maximum occlusion between agent and foreground objects in terms of 2D Intersection-over-Union (IoU).
Intuitively, this strategy generates challenging examples for the detector. We measured the average IoU between the 2D bounding box of the agent and all existing objects in each camera view, using this score to indicate the amount of occlusion caused after object placement. To avoid degenerative cases where the agent completely occludes other objects (e.g., placement very close to the camera), we penalized the occlusion score by the number of fully occluded objects across the search seeds. \cref{tab:ablation} reports the results of augmenting the entire nuScene dataset with these two variations in agent placement. The results indicate that incorporating highly occluded examples does not improve training in terms of $mAP$ measure. However, high occlusion samples improve the detector's performance in predicting object poses with achieving a lower mAOE score. 
\cref{fig:occlusion} visualizes two examples of augmented cameras with minimal and maximal occlusion, each by searching over 16 random seeds in the drivable space.

\section{Discussion and Conclusion}
This paper presents a straightforward approach for data augmentation to enhance 3D object detection in autonomous driving. 
We explored the performance of 3D scene reconstruction and geometrically accurate object insertion, comparing them to state-of-the-art diffusion methods, including recent strategy that employ conditioning on BEV maps and 3D layouts. 
Our experiments show that on both monocular and multi-camera setup, image augmentation through 3D Gaussian Splatting outperforms competitive diffusion-based methods.
Moreover, this study addresses several important questions regarding which factors are more important for augmentation.
Through experiments on large-scale data, we highlighted that geometric diversity of augmented data is more rewarding than photometric diversity.
Moreover, we observe that a high degree of pose diversity in synthetic objects benefits the detector in terms of performance in orientation estimation.
Finally, we explored the effectiveness of generating hard examples, both in terms of causing high detector loss and creating occlusions in the scene, which we found to be unrewarding strategies.
We encourage further exploration of these aspects by the community.

{
    \small
    \bibliographystyle{ieeenat_fullname}
    \bibliography{references}
}
\clearpage
\appendix
\onecolumn

\begin{center}
\textbf{\Large Gaussian Splatting is an Effective Data Generator for 3D Object Detection \\ Supplementary Material}
\end{center}

\setcounter{section}{1}

\subsection{3D asset generation pipeline}
\begin{figure*}[h!]
    \centering
    \includegraphics[width=.8\linewidth, trim={1cm, 3cm, 0cm, 3cm}, clip]{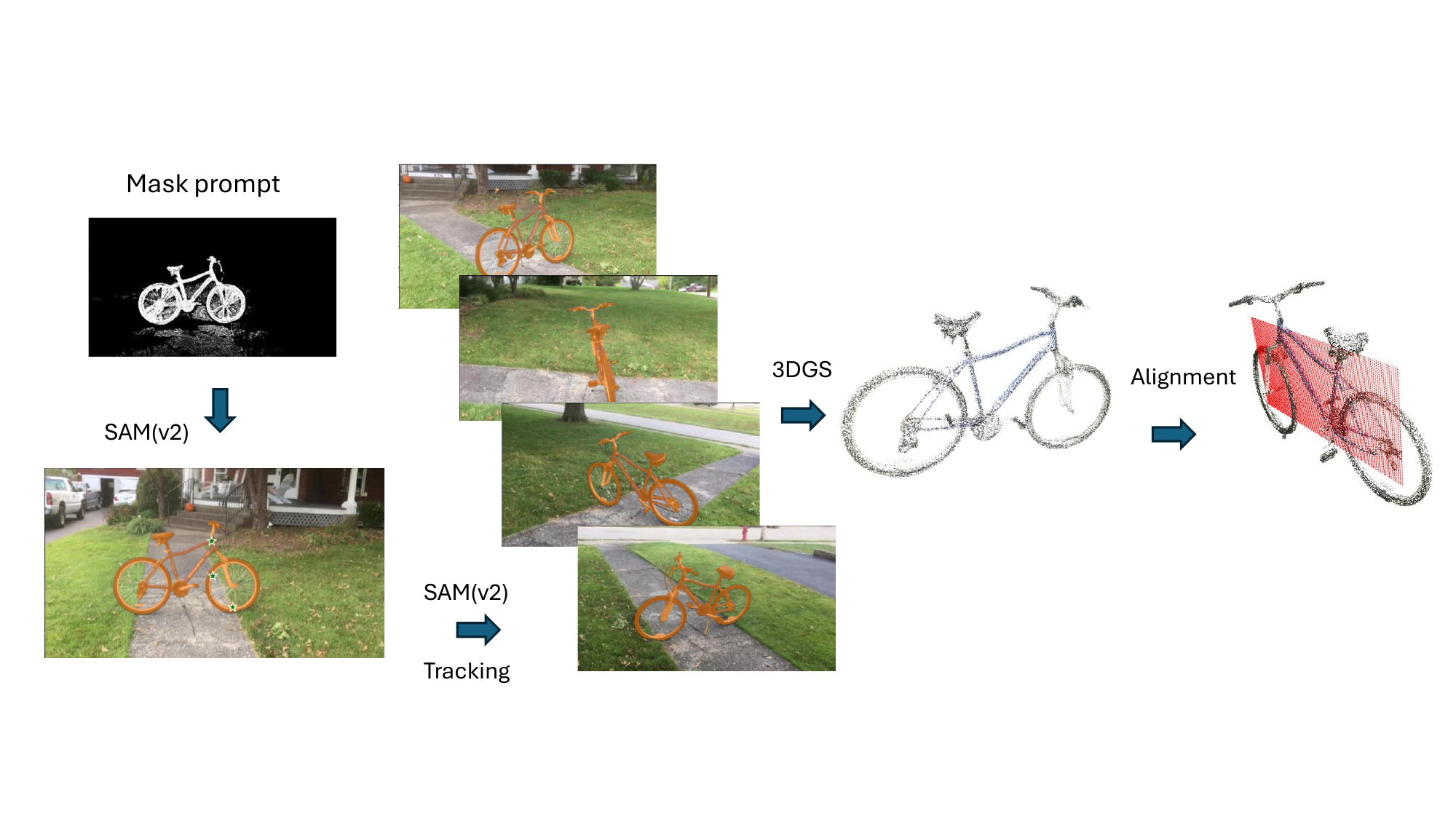}
    \caption{3D asset generation pipeline including accurate object mask generation, 3D reconstruction, and template-based pose alignment.}
    \label{fig:appendix:agent_gen}
\end{figure*}

\begin{figure*}[t]
 \resizebox{.99\linewidth}{!}{
\begin{tabular}{ll}

  \includegraphics[width=\linewidth, trim={5cm, 5cm, 4.3cm, 3.5cm}, clip]{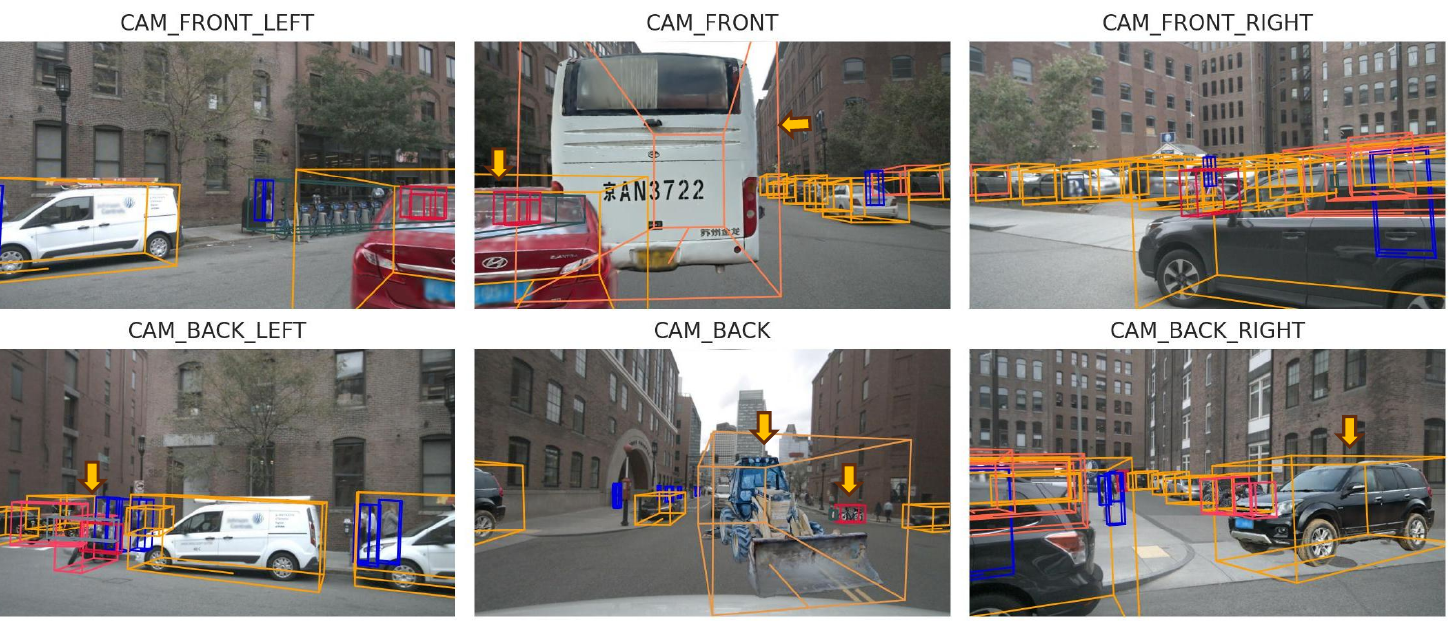}\\
  \midrule
  \includegraphics[width=\linewidth, trim={5cm, 5cm, 4.3cm, 4cm}, clip]{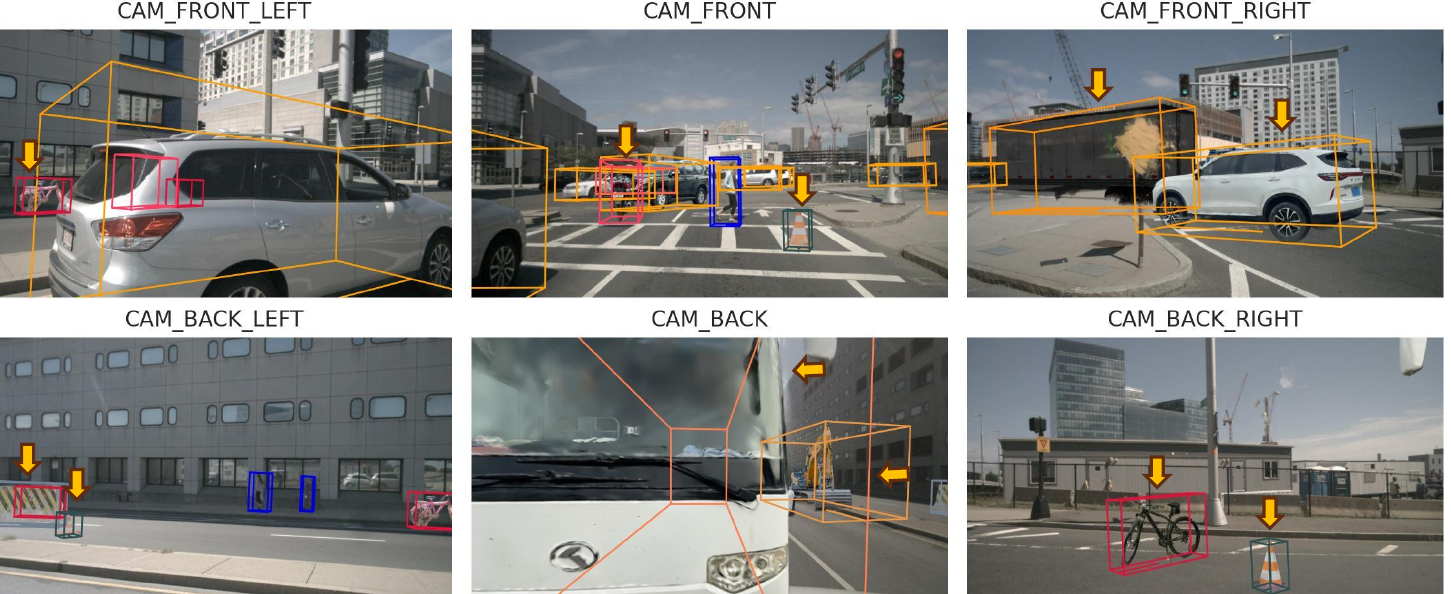}\\
  \midrule
  \includegraphics[width=\linewidth, trim={5.2cm, 5cm, 4.2cm, 4cm}, clip]{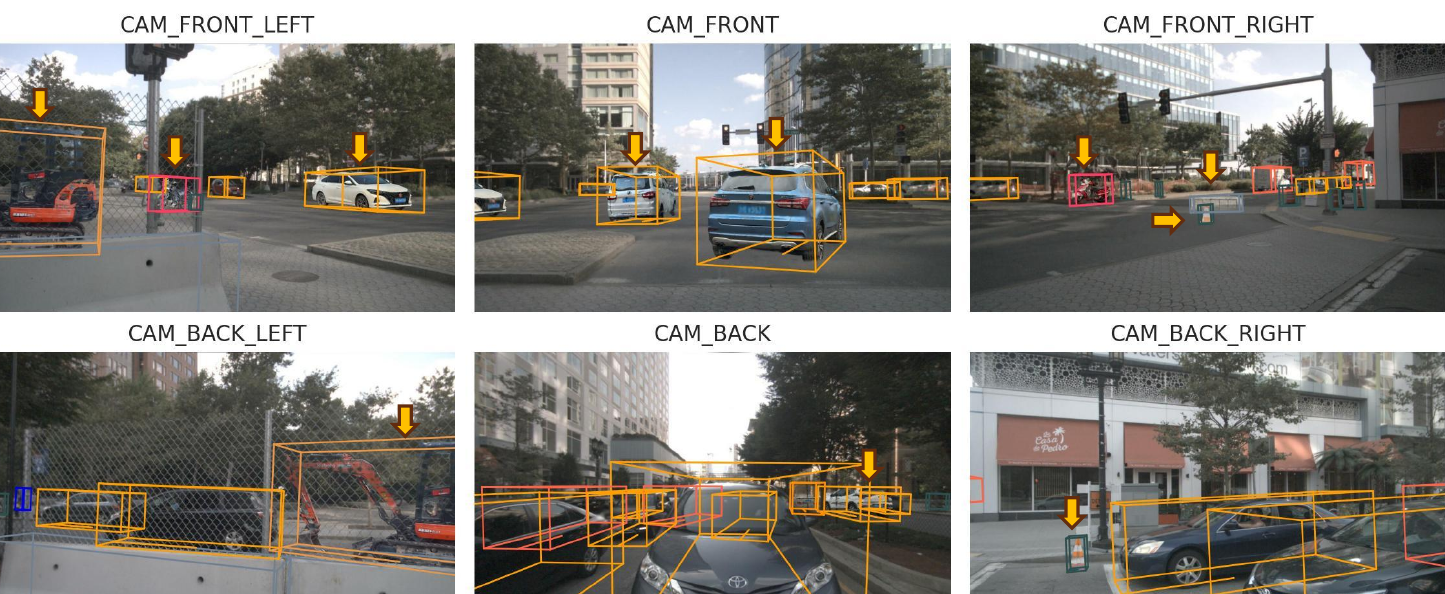}
  \end{tabular}
  }
  \caption{
     Additional visual examples of multi-camera 3D augmentation and their associated  3D bounding boxes. The highlighted arrows indicate the inserted agents.
  }
   \label{fig:3DGS_more_viz}
\end{figure*}

To generate 3D models of objects from multi-view image datasets, we first used SAM(v2) by prompting on a coarse 2D mask of the object in the initial image to obtain accurate 2D mask images of the target object across all multi-view images. We then used these images, along with the given camera poses and 2D masks, to generate 3D Gaussian models using the 3D-GS pipeline. The obtained 3D agents were reconstructed in their canonical coordinate system. Therefore, we employed a template-based matching procedure based on the Iterative closest point (ICP) algorithm to align them to a universal coordinate system. Figure~\ref{fig:appendix:agent_gen} shows the 2D mask generation and 3D Gaussian models for one of the objects from CO3D dataset, used for data augmentation.
. \cref{fig:3DGS_more_viz} visualizes some examples of augmented cameras using our method. 

\subsection{Depth-ControlNet baseline}

To augment the nuScenes \cite{caesar2020nuscenes} training dataset, we employ a ControlNet \cite{controlnet} with depth conditioning to replace existing vehicles.
This ``object replacement'' strategy is similar to prior work on object detector augmentation \cite{gen2det, kupyn2024dataset}, which shows that diversifying the appearance of objects without changing their location can be an effective augmentation approach, albeit for 2D task.

\begin{figure*}[t!]
    \centering
    \includegraphics[width=1.0\linewidth]{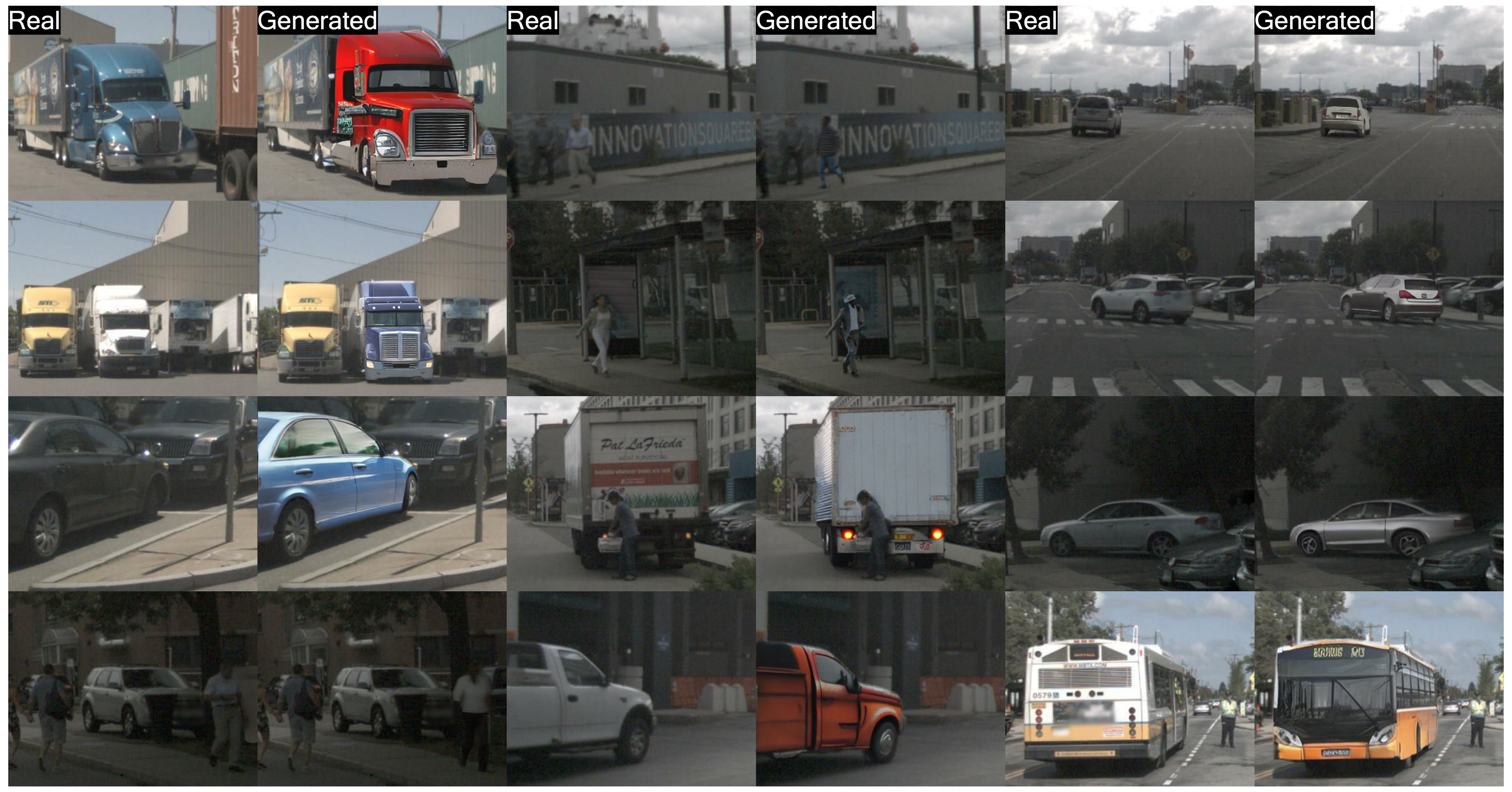}
    \caption{Examples of original nuScenes objects and their replacements, generated using the depth-ControlNet. This augmentation strategy replaces visible objects while keeping the background intact. Best viewed electronically.}
    \label{fig:appendix:depthcn_baseline}
\end{figure*}

We first filter the training data using nuScenes' visibility filters with a minimum of $3$, and select instances from all object categories. This leaves 67,203 object instances. We take square crops around the objects of interest of $1.5\times$ the size of the longest side of the bounding box, resize to $512\times512$, and obtain instance masks and depth maps for each crop using an off-the-shelf instance segmentation model \cite{kirillov2023segment} and depth estimator \cite{depthanything}.
To augment the training data we use a pre-trained \emph{StableDiffusionv1.5} checkpoint for the diffusion model (\texttt{runwayml/stable-diffusion-v1-5}) and the \texttt{lllyasviel/control\_v11f1p\_sd15\_depth} checkpoint for the ControlNet. Both checkpoints are publicly available from HuggingFace \cite{huggingface}. For each object, we take a square crop following training settings, mask the object location using the previously extracted instance mask, and perform inference with the DDIM \cite{song2020denoising} scheduler for 15 steps, with classifier-free guidance scale $7.5$. The resulting object crops are placed back into the original frame using the instance mask. If no object is present in the frame, we simply copy the frame. We thus effectively double the training set, adding one augmented image for every real image. We show example visualizations of original and augmented frames in Fig.~\ref{fig:appendix:depthcn_baseline}.

\subsection{MagicDrive baseline}
\begin{figure*}[t]
\begin{tabular}{ccc}
\includegraphics[width=0.3\linewidth]{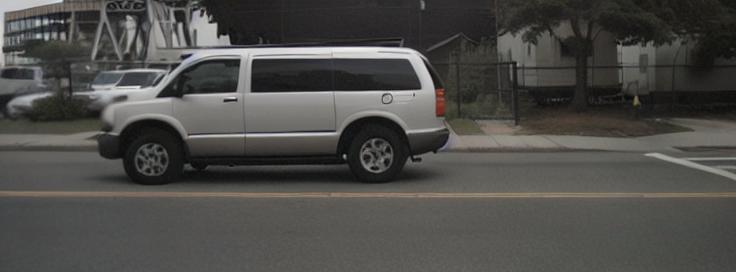}&
\includegraphics[width=0.3\linewidth]{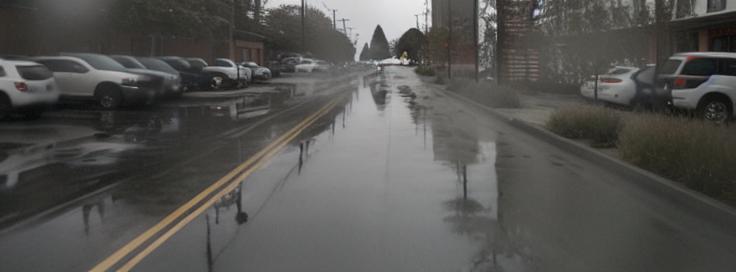}&
\includegraphics[width=0.3\linewidth]{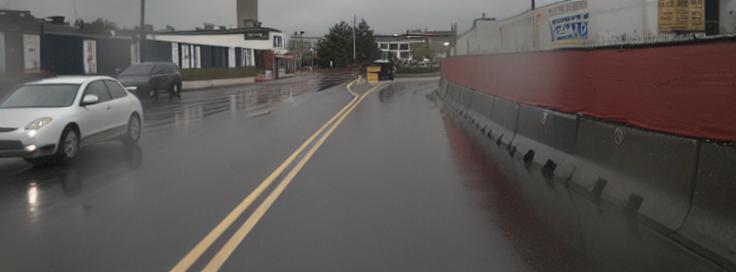} \\
\includegraphics[width=0.3\linewidth]{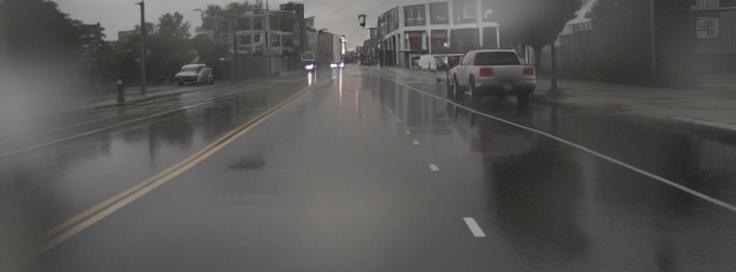}&
\includegraphics[width=0.3\linewidth]{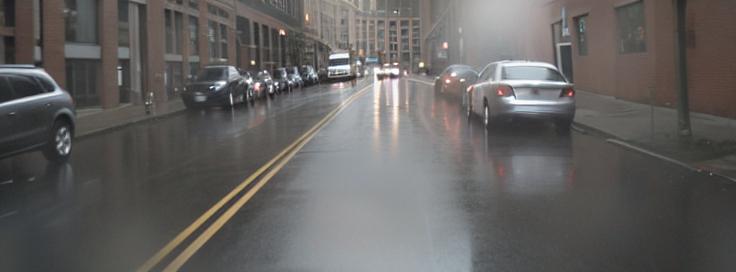}&
\includegraphics[width=0.3\linewidth]{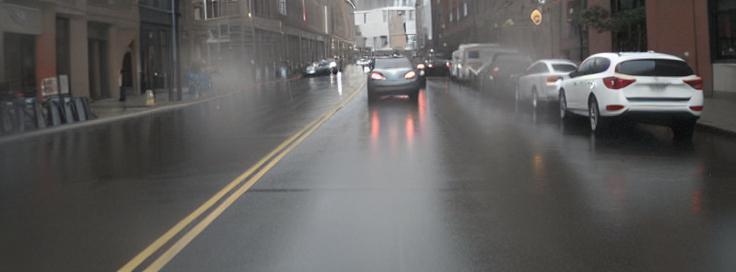}
\end{tabular}
\vspace{-5pt}
  \caption{Visual examples of $272 \times 736$ generated images (frontal camera) using MagicDrive~\cite{gao2023magicdrive}.}
  \label{fig:magicdrive_samples}
\end{figure*}

For the MagicDrive experiments, we utilized the author's released checkpoint\footnote{\url{https://github.com/cure-lab/MagicDrive}} to generate images with a resolution of 272$\times$736 pixels. We followed a procedure where 50\% of the bounding boxes were dropped in each camera frame, and images were generated conditioned on the remaining boxes. Alternatively, during inference, all bounding boxes were preserved, and images were generated with all ground truth boxes. For training 3D detectors and working with a mixture of real and augmented data, the images were zero-padded to match the original dimensions of the NuScenes samples. The evaluation was conducted on the NuScenes validation set using their original images. \cref{fig:magicdrive_samples} visualizes some generated images using MagicDrive model.

\begin{figure*}
    \centering
    \includegraphics[trim={0 0 10cm 0},clip,width=0.9\textwidth]{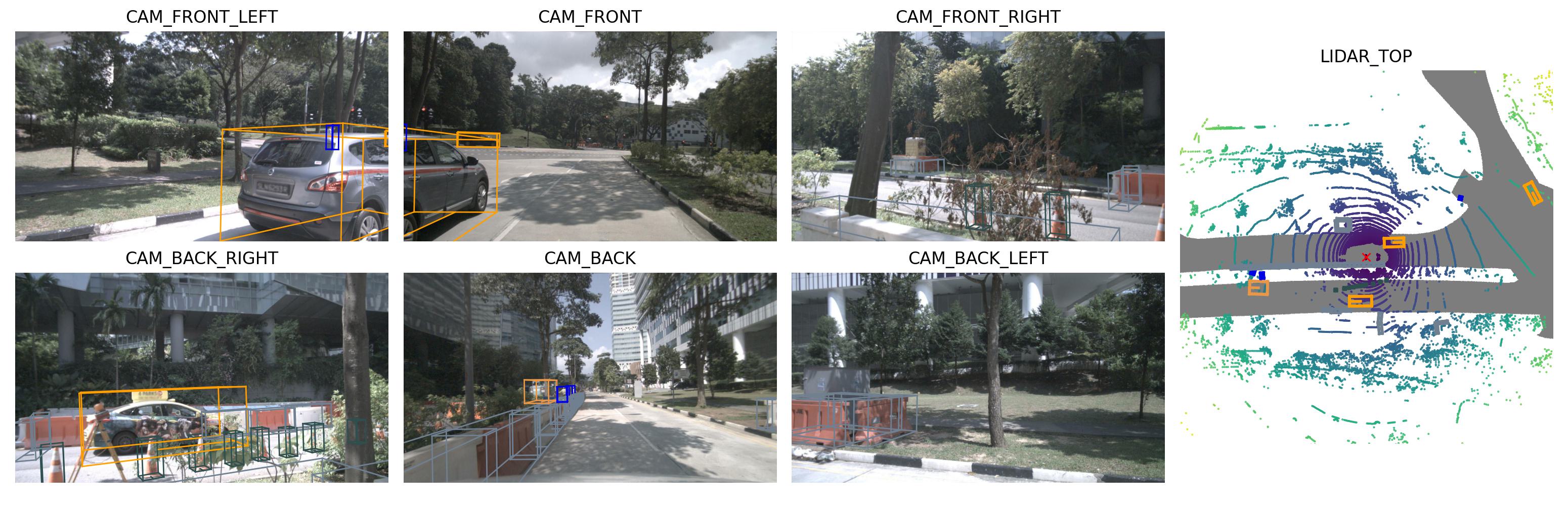}
    \includegraphics[trim={0 0 10cm 0},clip,width=0.9\textwidth]{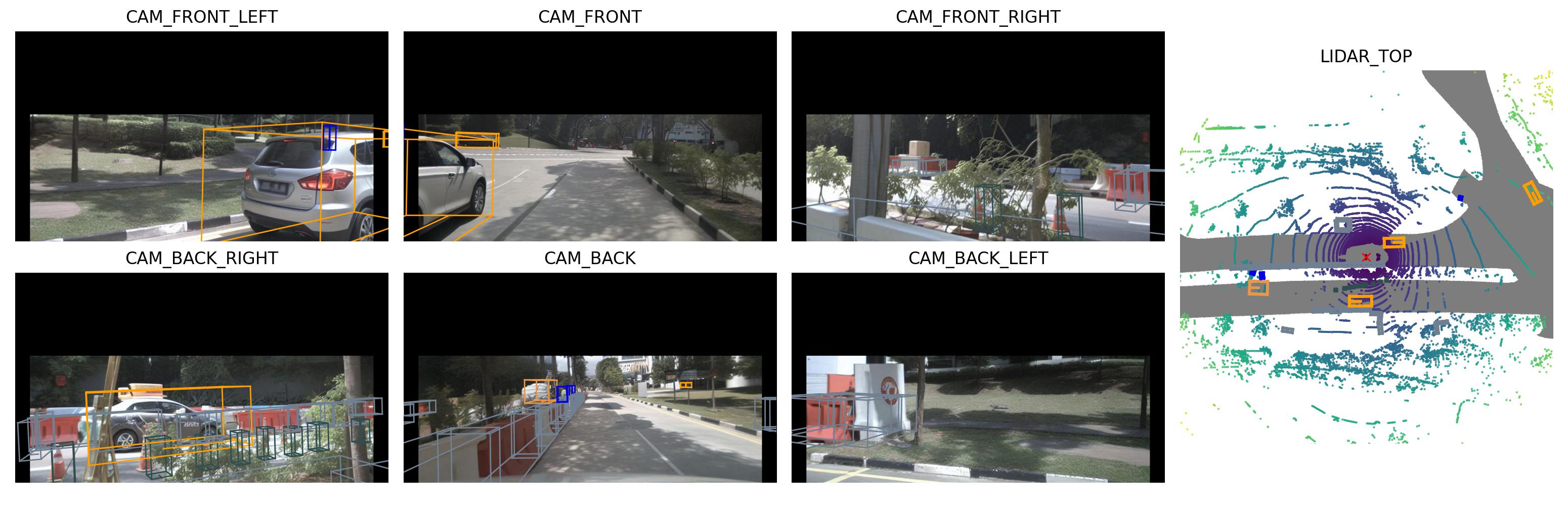}
    \includegraphics[trim={0 0 10cm 0},clip,width=0.9\textwidth]{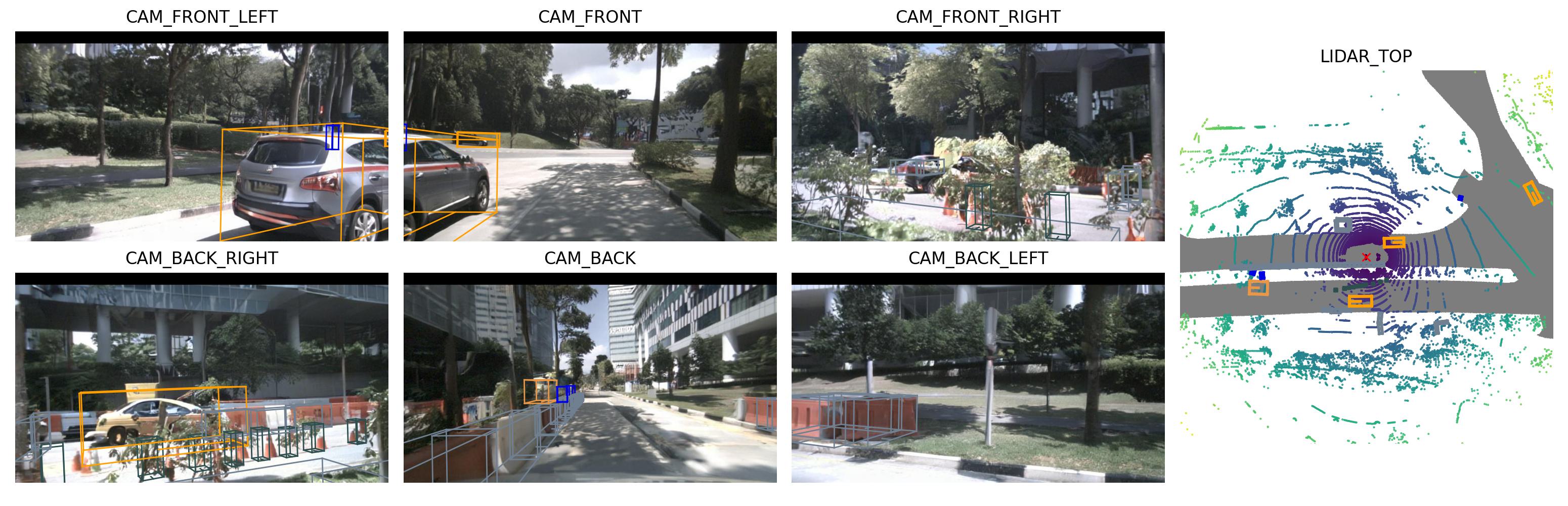}
    \caption{Example comparison of original frame (top), a padded MagicDrive generation (middle) and a MagicDriveDiT generation (bottom). 
    As we use the unmodified annotation as conditioning, generation results in highly similar frames.
    }
    \label{fig:appendix:magicdrive_comparison_1}
\end{figure*}

\begin{figure*}
    \centering
    \includegraphics[trim={0 0 10cm 0},clip,width=0.9\textwidth]{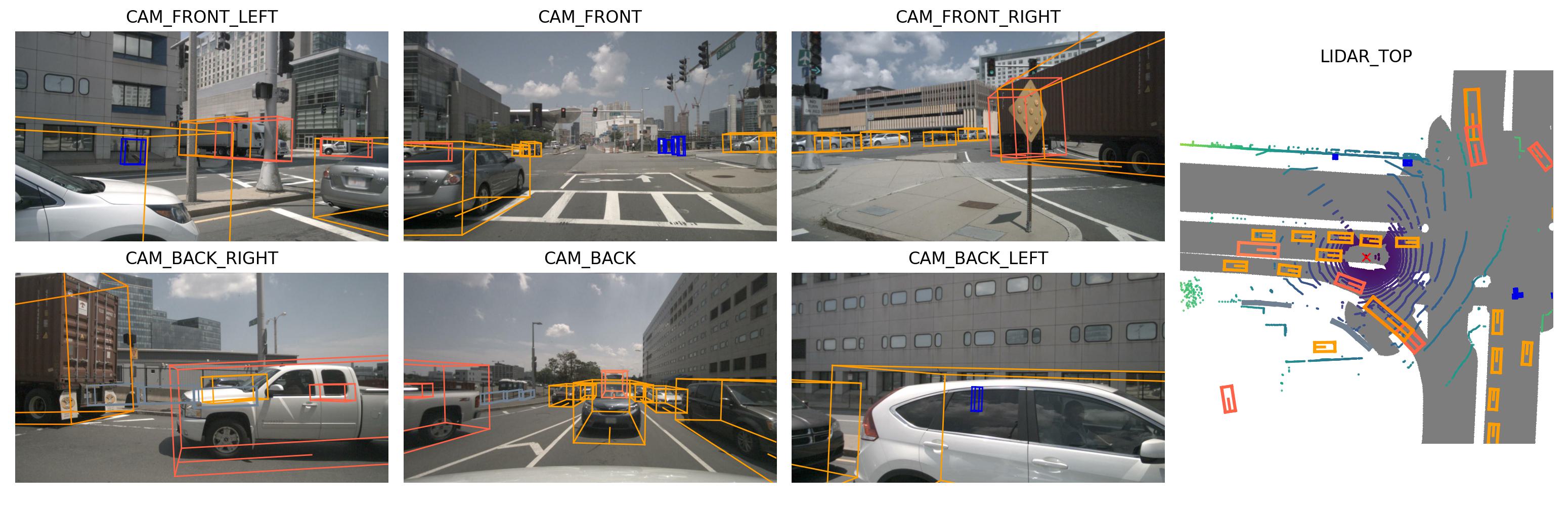}
    \includegraphics[trim={0 0 10cm 0},clip,width=0.9\textwidth]{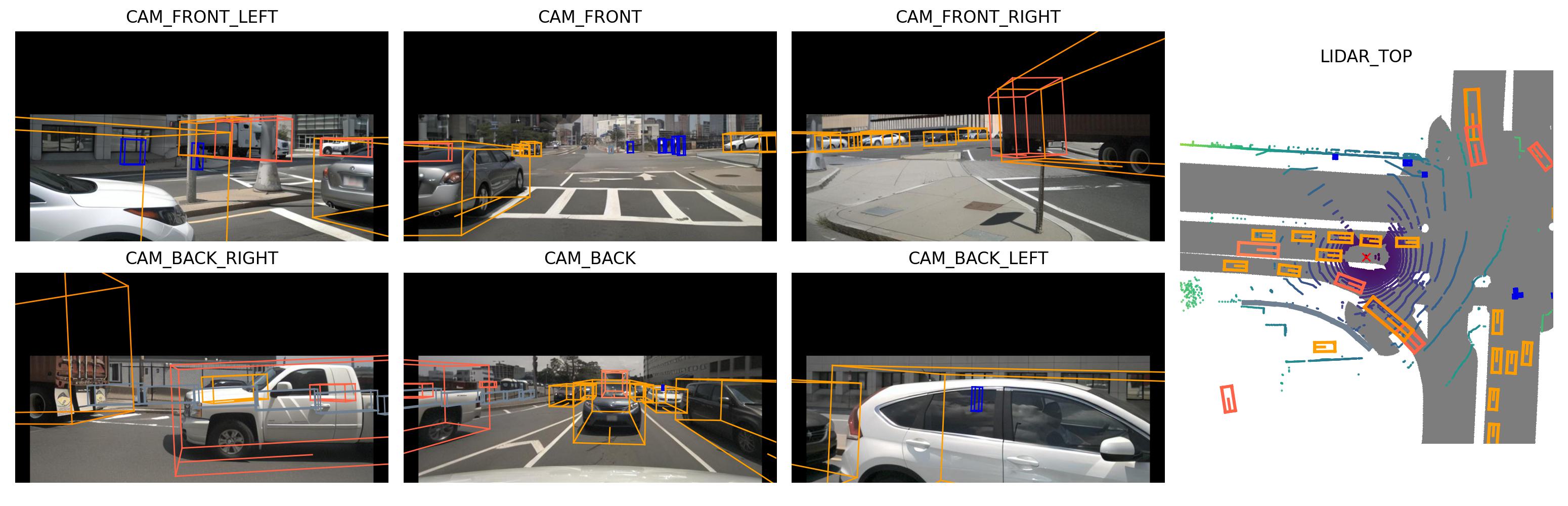}
    \includegraphics[trim={0 0 10cm 0},clip,width=0.9\textwidth]{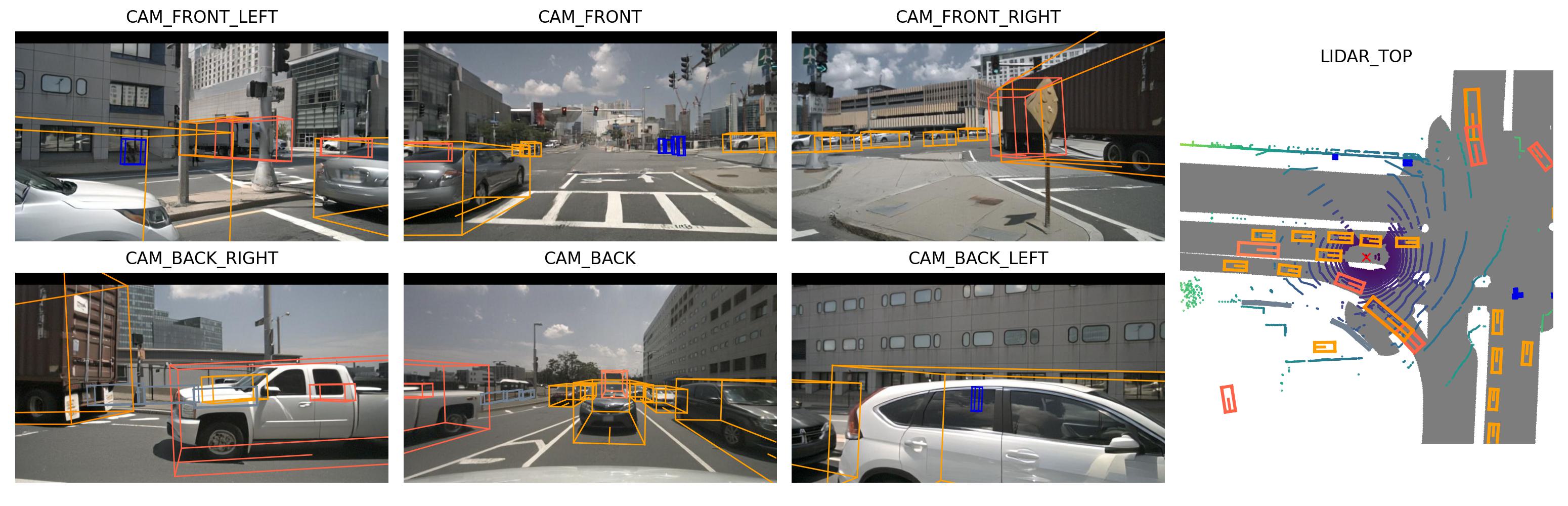}
    \caption{
    Example comparison of original frame (top), a padded MagicDrive generation (middle) and a MagicDriveDiT generation (bottom).
    As we use the unmodified annotation as conditioning, generation results in highly similar frames.
    }
    \label{fig:appendix:magicdrive_comparison_2}
\end{figure*}

\subsection{MagicDriveDiT baseline}
The MagicDriveDiT baseline follows a similar setup as the MagicDrive baseline, also using the officially released checkpoint\footnote{\url{https://github.com/flymin/MagicDriveDiT}}. The difference is that here we create images at a 424$\times$800 resolution. For evaluation, images are padded to the correct aspect ratio and then upsampled by a factor 2$\times$. In principle, MagicDriveDiT is capable of producing frames at quasi-full resolution of 848$\times$1600, but the compute requirements proved prohibitive for us. 
\cref{fig:appendix:magicdrive_comparison_1} and \cref{fig:appendix:magicdrive_comparison_2} visualize some generated images using both MagicDrive and MagicDriveDiT models.

\end{document}